\RequirePackage{fix-cm}
\documentclass[twocolumn]{svjour3}          %
\smartqed  %
\usepackage{graphicx}

\usepackage[textwidth=10em,textsize=small]{todonotes}
\usepackage{amsmath}
\usepackage{natbib}
\usepackage{xcolor,breqn}
\usepackage{hyperref}

\RequirePackage{fix-cm}
\usepackage{enumitem}
\usepackage{graphicx}
\usepackage{color}
\usepackage{times}
\usepackage{latexsym}
\usepackage{enumerate}
\usepackage{epsfig}
\usepackage{amsmath}
\usepackage{amssymb}
\usepackage{amsfonts}
\usepackage[english]{babel}
\usepackage{graphicx}
\usepackage{graphics}
\usepackage{psfrag}
\usepackage{amsbsy}
\usepackage{amsmath}
\usepackage{textcomp}
\usepackage{booktabs}
\usepackage{amssymb}
\usepackage{amsmath}
\usepackage{bm}
\usepackage{enumerate}
\usepackage{algorithm}
\usepackage{algpseudocode}
\usepackage{multirow}
\usepackage{verbatim}
\usepackage{caption}
\usepackage{subcaption}
\usepackage{tabularx}
\usepackage{listings}
\usepackage{soul}
\usepackage{pbox}

\def\be{\mathbf{e}}
\def\bx{\mathbf{x}}

\def\bv{\mathbf{v}}

\def\bF{\mathbf{F}}

\def\bX{\mathbf{X}}

\def\bU{\mathbf{U}}

\def\bZ{\mathbf{Z}}
\def\b0{\mathbf{0}}
\def\b1{\mathbf{1}}

\def\cN{\mathcal{N}}

\def\bmu{\mbox{\boldmath $\mu$}}

\def\btheta{\mbox{\boldmath $\theta$}}

\def\bSigma{\mbox{\boldmath $\Sigma$}}

\def\bLambda{\mbox{\boldmath $\Lambda$}}

\def\bPsi{\mbox{\boldmath $\Psi$}}
\def\bTheta{\mbox{\boldmath $\Theta$}}

\def\tr{\mathrm{tr}}

\DeclareMathOperator*{\argmax}{arg\,max}

\begin{document}

\title{Model-based Clustering using Automatic Differentiation: Confronting Misspecification and High-Dimensional Data 
}

\titlerunning{Clustering using Automatic Differentiation}        %

\author{Siva Rajesh Kasa$^*$\thanks{$^*$Corresponding Author} \and Vaibhav Rajan %
}

\authorrunning{Kasa and Rajan} %

\institute{ \at
              Department of Information Systems and Analytics\\School of Computing, National University of Singapore\\
              \email{kasa@u.nus.edu, vaibhav.rajan@nus.edu.sg}           %
}

\date{Received: date / Accepted: date}

\maketitle

\begin{abstract}
We study two practically important cases of model based clustering using Gaussian Mixture Models: (1) when there is misspecification and (2) on high dimensional data, in the light of recent advances in Gradient Descent (GD) based optimization using Automatic Differentiation (AD).
Our simulation studies show that EM has better clustering performance, measured by Adjusted Rand Index, compared to GD in cases of misspecification, whereas on high dimensional data GD outperforms EM.
We observe that both with EM and GD 
there are many solutions with high likelihood but poor cluster interpretation.
To address this problem we design a new penalty term for the likelihood based on the Kullback Leibler divergence between pairs of fitted components.
Closed form expressions for the gradients of this penalized likelihood are difficult to derive but AD can be done effortlessly, illustrating the advantage of AD-based optimization.
Extensions of this penalty for high dimensional data and for model selection are discussed.
Numerical experiments on synthetic and real datasets demonstrate the efficacy of clustering using the proposed penalized likelihood approach.

\end{abstract}

\section{Introduction}\label{intro}

Model-based clustering is a well-established paradigm for clustering multivariate data
\citep{banfield1993model,melnykov2010finite}.
In this approach, data is assumed to be generated by a finite mixture model where each component represents a cluster.
For continuous-valued variables, it is common to model each component density by a multivariate Gaussian distribution.
The rigorous probabilistic framework of mixture models facilitates model building and assessment. Gaussian Mixture Models (GMM), in particular,  are widely used in a variety of applications \citep{McLa:Peel:fini:2000}.
Expectation Maximization (EM) and its variants are by far the most popular methods to obtain Maximum Likelihood Estimates (MLE) 
of the parameters of a GMM \citep{dempster1977maximum,mclachlan2007algorithm}.

The one-to-one correspondence between fitted components and clusters, that makes model-based clustering intuitive and appealing, assumes that the underlying  model is correctly specified and each data cluster can be viewed as a sample from a mixture component.
In real data, the true distribution is rarely known and further,
data may be contaminated by outliers from a distribution different from the assumed model.
In such misspecified settings, MLE may be unrealistic and may fail to recover the underlying cluster structure \citep{farcomeni2016robust}.
Another related problem %
is that of spurious local maximizers.
It is well known that the likelihood of GMM with unrestricted covariance matrices is unbounded \citep{day1969estimating}.
As a result, one finds solutions from EM with fitted `degenerate' component(s) having very small variance 
corresponding to a cluster containing very few closely located data points; in the case of multivariate data, there are components with very small generalized variance, lying in a lower-dimensional subspace and close to the boundary of the parameter space
\citep{Peel:McLa:2000}.
They may have equal or higher likelihood compared to other solutions but without adequate interpretability in terms of cluster structure.
Various constraints on the parameter space have been proposed to obtain non-spurious solutions through EM \citep{ingrassia2004likelihood,Ingr:Rocc:2007,Chen2009,ingrassia2011degeneracy}.
Approaches to fit contaminated mixtures include trimming and use of constraints \citep{cuesta1997trimmed,ruwet2013breakdown,garcia2014constrained}
and parsimonious models \citep{punzo2016parsimonious}.
Recent theoretical results on misspecification in  EM-based inference can be found in \citep{dwivedi2018theoretical}. 

Model-based clustering is also challenging 
for high dimensional data, particularly when the
number of observations is relatively low.
When the number of observations is small compared to the total number of free parameters to be estimated, the estimated covariance matrices are ill-conditioned and leads to unstable clustering \citep{bouveyron2014model}.
When the number of observations is less than the data dimensionality, EM fails
because the E-step requires the inversion of a matrix, that is typically not full-rank in such cases, and there are singularities in the parameter manifolds \citep{amari2006singularities}.
Computational difficulties also arise due to quadratic increase in covariance matrix parameters with increasing dimensions.
Common solutions include combining clustering with local dimensionality reduction, e.g., in the Mixture of Factor Analyzers (MFA)
\citep{Ghah:Hilt:1997} and its extensions such as Parsimonious GMMs \citep{McNi:Murp:Pars:2008} that impose constraints on the covariance structures.
MLE for MFA and related models are also prone to spurious solutions and singularities \citep{greselin2015maximum,garcia2016joint}.
Further, constraints on the covariance may lead to incorrect orientations or shapes of the resulting clusters \citep{zhou2009penalized}. 
A recent method by \cite{cai2019chime}, uses a modified E-step  that obviates the need to compute the inverse of the covariance matrix but
it assumes equi-covariance
in all clusters and requires solving an expensive quadratic optimization problem.
Another class of EM-based approaches are variable selection methods \citep{fop2018variable}. 
Many of these approaches either fail or do not yield satisfactory clusters in high-dimensional low sample size regimes \citep{kasa2019gaussian}.
Further, in high dimensions likelihood-based criteria 
for model selection also fail \citep{giraud2014introduction}.

Gradient-based methods, as alternatives to EM, for MLE of GMMs have been studied previously \citep{redner1984mixture,
jordanEMasGD,alexandrovich2014exact}.
Hybrid methods have also been proposed \citep{atkinson1992performance,jamshidian1997acceleration}.
In general, methods based on conjugate gradients, quasi-Newton or Newton have been found to be inferior to EM \citep{jordanEMasGD}.
These studies do not investigate cases of misspecification or high-dimensional settings.
Our present work leverages 
Automatic Differentiation (AD) tools, 
that obviate the need to derive closed-form expressions of gradients and thereby facilitate Gradient Descent (GD) based inference\footnote{Most AD solvers use minimization as the canonical problem and ML estimation is done by Gradient Descent on the negative loglikelihood. Throughout the paper, we use the term {\it Gradient Descent} for maximization problems as well, instead of Gradient Ascent, with the assumption that the sign of the objective function is changed during optimization.}
 in complex models such as MFA.
The computational efficiency and numerical accuracy offered by AD has been successfully leveraged in training over parameterized deep neural networks.
Few previous works have investigated AD for optimization in statistical models -- some examples include %
non-linear random effects models \citep{fournier2012ad,skaug2006automatic,skaug2002automatic} and Bayesian models \citep{kucukelbir2017automatic}.
AD-based Gradient Descent (AD-GD) for mixture models has been discussed previously in \cite{maclaurin2015autograd}. 
However, to our knowledge, there is no previous work comparing the clustering performance of AD-GD with EM in the context of misspecification and high dimensional data, which is the focus of this paper.

Our extensive empirical analysis on simulated data shows that, in general, EM has better clustering performance (in terms of Adjusted Rand Index (ARI)) in cases of misspecification while AD-GD outperforms EM on high-dimensional data.
However, we observe that,  in misspecified cases, both EM and AD-GD can lead to poor clusterings for many different initializations.
Similar to spurious solutions, these clusterings have high likelihood and low ARI, but they differ in other characteristics.
For instance, we find that these solutions have components with large variance and occur frequently with many different initializations.
With high dimensional data, we find that the use of AD-GD leads to components in the fitted GMM with low variance and high mixture weight  (more details are in \S \ref{sec:autodiff}).
To distinguish them from solutions previously characterized as spurious solutions, we call them {\it inferior} clusterings.
To address the problem of inferior clusterings, we propose a new penalty term based on the Kullback Leibler divergence between pairs of fitted component Gaussians.
Closed form expressions of the gradients for this penalty term are difficult to derive, but AD-based optimization can be done effortlessly.
Extensions of this penalty for high-dimensional data and for model selection are discussed.
Experiments on synthetic and real datasets demonstrate the efficacy of this penalized likelihood for cases of misspecification and high-dimensional data.

This paper is organized as follows.
We begin with a brief overview of Mixture Models and Gradient Descent based inference using Automatic Differentiation (\S \ref{sec:background}).
We present our empirical study on clustering performance of EM and AD-GD on misspecified cases and high-dimensional data (\S \ref{sec:autodiff}).
Our proposed penalty term is then introduced (\S \ref{sec:penalty})
and the following section (\S \ref{sec:SIA}) describes our clustering algorithms using this penalized likelihood.
Next, we present the results of our simulation studies (\S \ref{sec:sim}). 
We discuss our experiments on real datasets (\S \ref{sec:real}) and then conclude (\S \ref{sec:concl}).

\section{Background}
\label{sec:background}

\subsection{Mixture Models}

Let $f(\boldsymbol{x} ; \boldsymbol{\btheta})$ be the density of a $K$-component mixture model.
Let $f_k$ denote the $k^{\rm th}$ component density with parameters  $\btheta_k$ and weight  $\pi_k$.
The density of the mixture model is given by
$$f(\boldsymbol{x} ; \boldsymbol{\btheta}) =\sum_{k=1}^{K} \pi_{k} f_{k}\left(\boldsymbol{x} ; \boldsymbol{\btheta}_{k}\right), $$
where  $\sum_{k=1}^{K} \pi_k = 1$ and $\pi_k \ge 0$ for  $k = 1,\ldots, K$ and
$\btheta$ %
denotes the complete set of parameters of the model. %
In a GMM, each individual component $f_k$ is modeled using a multivariate Gaussian distribution $\mathcal{N}(\bmu_k, \bSigma_k)$ where $\bmu_k$ and $\bSigma_k$ are its mean and covariance respectively. 
Appendix \ref{sec:symbols} has a list of symbols and notation used in the paper for reference.

Given $n$ independent and identically distributed  $(iid)$
instances of $p$-dimensional data, $[x_{ij}]_{n \times p} $
where index
$i$ is used for observation, and $j$ is used for dimension.
Maximum Likelihood Estimation (MLE) 
aims to find parameter estimates
$\hat \btheta $ from the overall parameter space $\bTheta$ of $f(\btheta)$ such that probability of observing the data %
samples $\bx_1, \dots, \bx_n$ from this family 
is maximized, i.e.,
 $\hat{\btheta}=\argmax_{\btheta \in \bTheta} \mathcal{L}(\btheta)$, where, 
 $\mathcal{L(\btheta)} = \frac{1}{n}\sum_i \log f(\bx_i;\btheta) $ is the loglikelihood.

Following \citet{white1982maximum}, if the observed data 
are $n$  $iid$ samples  from a probability distribution                        
$P(\eta^*)$ (where 
$\eta^*$ denotes the \textit{true} set of parameters)
and the fitted model has the same
functional form $P(.)$, then the model is said to be correctly specified.
Otherwise,
the model is said to be \textit{misspecified}.
An example of misspecification is when the data is sampled from a multivariate $t$-distribution, and clustered using a GMM. It is easy to see that many real-world applications of model-based clustering are prone to misspecification.

\subsection{Mixture of Factor Analyzers (MFA)}
Without any restrictions, the GMM is a highly parameterized model where
$K-1$ mixture weights, $Kp$ elements of the mean vectors and $\frac12 Kp(p+1)$ covariance matrix elements have to be estimated which can be difficult in high dimensions.
MFA provides a parsimonious characterization through local dimensionality reduction using $q$-dimensional latent factors $\bF_{i}$ that model correlations in the data, where $q < p$ \citep{Ghah:Hilt:1997}.
Each observation $\bx_i$ (for $i=1,\ldots,n$) is assumed to be generated as follows:
\begin{align}
\bx_i=\bmu_k+\bLambda_k\bF_{ik}+\be_{ik} \quad\textrm{with probability } \pi_k  
\label{factor_an}
\end{align}
where $\bLambda_k$ is a $p \times q$ matrix of \textit{factor loadings} and both the \textit{factors} 
$\bF_{ik}$ and \textit{errors} $\be_{ik}$ follow Gaussian distributions that are independent of each other:
\begin{align*}
\bF_{1k},\ldots, \bF_{nk} \sim \cN(\mathbf{0},\mathbb{I}_q); \quad \quad
\be_{ik} \sim \cN(\mathbf{0},\bPsi_k)
\end{align*}
where $\mathbb{I}_q$ is the $q$-dimensional Identity matrix and $\bPsi_k$ is a $p \times p$ diagonal matrix ($k = 1,\ldots,K$). 
Thus, the component covariance matrices have the form $\bSigma_k = \bLambda_k \bLambda_k^' + \bPsi_k$.
Since $\frac12 q(q-1)$ constraints are required to uniquely define $ \bLambda_k$, the 
 number of free parameters to be estimated is 
$ K(pq - \frac12 q(q-1) + p)$. 
See, e.g., \cite{greselin2015maximum} for details.

\subsection{Model Selection} 
Criterion-based model selection methods are designed to favor parsimony in modeling, by penalizing overfitting of the model. 
Common likelihood-based criteria include
Akaike Information Criterion (AIC):
$$\text{AIC} := 2p_e - 2\mathcal{L(\hat{\btheta})},$$
and Bayesian Information Criterion (BIC): 
$$\text{BIC} :=  p_e\ln(n) - 2\mathcal{L(\hat{\btheta})},$$
where $p_e$ is the number of parameters to be estimated in the model. 

It has been pointed out that MLE based criteria such as AIC and BIC are based on asymptotic normality and hence, cannot be used for model selection in high-dimensions where $n << p$ \citep{amari2006singularities, akaho2000nonmonotonic, giraud2014introduction}. 
Moreover, in high dimensions, it has been observed that BIC tends to underestimate the number of components  whereas AIC overestimates the number of components \citep{melnykov2010finite}.  

\subsection{Spurious Solutions}

Spurious solutions are local maximizers of the likelihood function but lack real-life interpretability and hence do not provide a good clustering of the data.
It is a consequence of the unboundedness of the GMM likelihood function for unrestricted component covariance matrices.
As discussed in \cite{McLa:Peel:fini:2000}, spurious solutions may be obtained when:
\begin{itemize}
\item 
a fitted component has very small non-zero variance for univariate data or generalized covariance, i.e., the determinant of the covariance matrix, for multivariate data. Such a component corresponds to a cluster containing very few data points close together or, for multivariate data, in a lower dimensional subspace.
\item
the model fits a small localized random pattern in the data instead of an underlying group structure.  Such solutions have very few points in one cluster with little variation, compared to other clusters, in the cluster's axes, or for multivariate data, small eigenvalues for the component covariance matrix.
\item
where the likelihood increases by fitting the covariance matrix of a component on just one or a few datapoints distant from the other samples.
\end{itemize}

In some cases, e.g., when a component is fitted over very few datapoints, spurious solutions lead to singularities in the component covariance matrices that can be detected during EM inference.
There are other cases as well, when the parameters lie close to the boundary of the parameter space, when the component covariance matrices are not singular but may be close to singular for some components.

\cite{McLa:Peel:fini:2000} observed that convergence to spurious solutions happen rarely; they are dependent on initialization -- often occurring only for some initializations; and convergence to non-spurious global maxima becomes difficult with increasing dimensionality.

Both noise and outliers can lead to 
spurious 
solutions.
Contaminating outliers from a population different from the assumed model are called mild outliers  \citep{ritter2014robust}).
If additional procedures, such as trimming \citep{cuesta1997trimmed, garcia2010review} are not performed, then fitting a GMM leads to misspecification.

\subsection{Gradient Descent and Automatic Differentiation}
In traditional GD based inference, gradients are required in closed form which becomes laborious or intractable to derive as the model complexity increases, e.g., for MFA. 
To automate the computation of derivatives three classes of techniques have been developed:
(a) Finite Differentiation (FD) (b) Symbolic Differentiation (SD) and (c) Automatic or Algorithmic Differentiation (AD).
Although easy to code, FD is slow at high dimensions and susceptible to floating point errors.
SD provides exact symbolic expressions of derivatives but has high computational complexity, in both time and memory and cannot be used when functions are defined using programmatic constructs such as conditions and loops.
The complexity and errors in both FD and SD increase for the computation of higher derivatives and partial derivatives of vector-valued functions.
AD overcomes all these limitations of FD and SD.
It provides efficient and accurate numerical evaluation of 
derivatives without requiring closed form expressions.

The numerical computation of a function can be decomposed into a finite set of elementary operations. 
These operations most commonly include arithmetic operations and transcendental function evaluations.
The key idea of AD is to numerically compute the derivative of a function by combining the derivatives of the elementary operations through the systematic application of the chain rule of differential calculus.
The efficiency of the computation is improved by storing evaluated values of intermediate sub-expressions that may be re-used.
Backpropagation, used to train neural networks, is a specific form of AD which is more widely applicable.
We refer the reader to recent surveys \citep{baydin2018automatic,margossian2019review} for more details. 
Appendix \ref{section:AD} discusses some examples.

Efficient implementations of AD are available in several programming languages and frameworks, e.g. Python \citep{maclaurin2015autograd}, R \citep{pav2016madness}, Pytorch \citep{paszke2017automatic} and Stan \citep{carpenter2015stan}.
Many first and second order gradient-based optimization algorithms are implemented in these libraries.
In this paper, we use Adam, a first order method that computes individual adaptive learning rates for different parameters from estimates of first and second moments of the gradients \citep{kingma2014adam}.

\section{Automatic Differentiation for Mixture Model based Clustering}
\label{sec:autodiff}

In this section we first summarize how Automatic Differentiation based Gradient Descent (AD-GD) can be used for GMM based clustering.
Second, we discuss our numerical studies that compare the performance of EM and AD-GD in cases of misspecification and high-dimensions.
Third, we explain the problems arising in GMM based clustering that are unresolved by both AD-GD and EM, which motivate the development of our new algorithm in the following section.

\subsection{Automatic Differentiation based Gradient Descent (AD-GD)}
\label{app:reparametrization}

To obtain MLE of GMMs EM elegantly solves 3 problems:
(1) Intractability of evaluating the closed-forms of the derivatives,
(2) Ensuring positive definiteness (PD) of the covariance estimates $\hat{\bSigma}_k$, and
(3) Ensuring the constraint on the component weights ($\sum_k \hat{\pi}_k = 1$ ).

In AD-GD, \textit{Problem 1} is solved inherently because we do not need to express the gradients in closed form by virtue of using AD. 
Further, the second-order Hessian matrix may also be evaluated using AD, 
enabling us to use methods with faster convergence.
To tackle \textit{Problem 2}, instead of gradients with respect to $\bSigma_k$, we compute the gradients with respect to $\bU_k$, where $\bSigma_k = \bU_k\bU^{T}_k$. 
We first initialize $\bU_k$ as identity matrices. Thereafter, we keep adding the gradients to the previous estimates of $\hat{\bU}^{t+1}_k$, i.e. 
\begin{align}
    \hat{\bU}^{t+1}_k  := \hat{\bU}^{t}_k + \epsilon  \frac{\partial \mathcal{L}}{\partial \bU_k} ; \, \;
    \hat{\bSigma}^{t+1}_k := \hat{\bU}^{t+1}_k \hat{\bU}^{{t+1}^{T}}_k 
\end{align}
where $\epsilon$ is the learning rate and superscripts $t, t+1$ denote iterations in GD. 
If the gradients are evaluated with respect to $\bSigma_k$ directly, there is no guarantee that updated $\hat{\bSigma}^{t+1}_k = \hat{\bSigma}^{t}_k + \epsilon \frac{\partial \mathcal{L}}{\partial \bSigma_k}$ will still remain PD. 
However, if the gradients are evaluated with respect to $\bU_k$, 
by construction $\hat{\bSigma}^{t+1}_k$ will always remains PD. 
Cholesky decomposition for reparameterizing $\bSigma_k$ can also be used \citep{salakhutdinov2003optimization}. 
In the case of MFA models, the component covariance matrices are in factorized form, 
$\bSigma_k = \bLambda_k \bLambda_k^T + \bPsi_k$, which ensures PD covariance matrix estimates.
\textit{Problem 3} is solved by using the log-sum-exp trick \citep{robert2014machine}. We start with unbounded $\alpha_k$ as the log-proportions: $\log \pi_k = \alpha_k - {\log(\sum_{k^'=1}^K e^{\alpha_{k^'}})}$.  We need not impose any constraints on $\alpha_k$ as the final computation of $\pi_k$ automatically leads to normalization, because $\pi_k = \frac{e^{\alpha_k}  }{\sum_{k^'=1}^K e^{\alpha_{k^'}}}$. Therefore, we reparametrize a constrained optimization problem into an unconstrained one (without using Lagrange multipliers) and we can update $\hat{\pi}_k$ as follows: %
\begin{align}
    \hat{\alpha}^{t+1}_k := \hat{\alpha}^{t}_k + \epsilon \frac{\partial \mathcal{L}}{\partial \alpha_k} ; \, \;
    \hat{\pi}^{t+1}_k := \frac{e^ {\hat{\alpha}^{t+1}_k}}{\sum_{k^'=1}^K e^{\hat{\alpha}^{t+1}_{k^'}}}
\end{align} 
Algorithm \ref{ALGO:AD-GD} shows all the steps for AD-GD inference of GMM and MFA.

\begin{algorithm}[t!]
\caption{AD-GD}
\label{ALGO:AD-GD}

\textbf{Input: }
Data: $n \times p$ dimensional matrix, 
required number of clusters $K$, learning rate $\epsilon$ and convergence tolerance $\gamma$.

\textbf{Initialize at iteration $t=0$:} 
$\hat{\bmu}^{0}_k, \hat{\alpha}_k$'s using K-Means or random initialization; 
$\hat{\bU}^{0}_k$ (for GMM) or $\hat{\bLambda}^0_k, \hat{\bPsi}^0_k$ (for MFA) as identity matrices.

\textbf{Objective:} $\text{argmax}_{\btheta}  \mathcal{L(\btheta)} $

\textbf{REPEAT:} At every iteration $t+1$: 

\resizebox{\linewidth}{!}{
  \begin{minipage}{\linewidth}
  \textbf{Mixture weight and mean updates}  
  \begin{align*}
 \hat{\alpha}^{t+1}_k := \hat{\alpha}^{t}_k + \epsilon \frac{\partial  \mathcal{L} }{\partial \alpha_k} ; \;
    \hat{\pi}^{t+1}_k  := \frac{e^{\hat{\alpha}^{t+1}_k} }{\sum_{k^'} e^{\hat{\alpha}^{t+1}_{k^'}}} ; \;
    \hat{\bmu}^{t+1}_k := \hat{\bmu}^{t+1}_k + \epsilon \frac{\partial  \mathcal{L}}{\partial \bmu_k}; \;
\end{align*}
  \end{minipage}
}

\resizebox{\linewidth}{!}{
  \begin{minipage}{\linewidth}
\textbf{Covariance updates}  
  
\textbf{-- For GMM:}
  \begin{align*}
    \hat{\bU}^{t+1}_k := \hat{\bU}^{t}_k + \epsilon \frac{\partial  \mathcal{L}}{\partial \bU_k} ; \;
    \hat{\bSigma}^{t+1}_k := \hat{\bU}^{t+1}_k \hat{\bU}^{{t+1}^T}_k 
    \end{align*}
 
\textbf{-- For MFA:}   
    \begin{align*}
    \hat{\bLambda}^{t+1}_k  := \hat{\bLambda}^{t}_k + \epsilon \frac{\partial \mathcal{L}}{\partial \bLambda_k}; 
    \hat{\bPsi}^{t+1}_k := \hat{\bPsi}^{t}_k + \epsilon  \frac{\partial \mathcal{L}}{\partial \bPsi_k}; 
    \hat{\bSigma}^{t+1}_k := \hat{\bLambda}^{t+1}_k \hat{\bLambda}^{{t+1}^{T}}_k + \hat{\bPsi}^{t+1}_k
\end{align*}
  \end{minipage}
}

\textbf{UNTIL:} convergence criterion $|\mathcal{L}^{t+1} - \mathcal{L}^{t} | < \gamma $ is met 

\end{algorithm}

\subsection{Misspecification}
\label{sec:misspec}
\begin{figure}[!h]
    \centering    
    \begin{minipage}{0.5\textwidth}
        \centering
                
\begin{subfigure}[b]{0.45\textwidth}
\includegraphics[width=\textwidth]{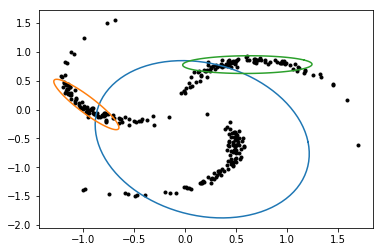}
        \caption{Set 1 - (0.625,\textbf{771.94})}
     \end{subfigure}        
~
\begin{subfigure}[b]{0.45\textwidth}
\includegraphics[width= \textwidth]{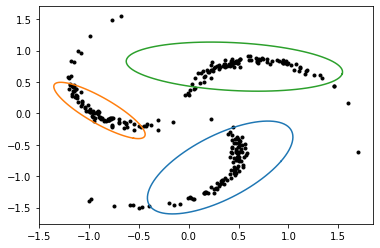}
        \caption{Set 2 - (0.912,781.05)}
     \end{subfigure}%

\begin{subfigure}[b]{0.45\textwidth}
\includegraphics[width=\textwidth]{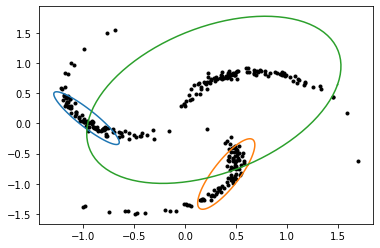}
        \caption{Set 3 - (0.651,786.82)}
    \end{subfigure}
    ~    
 \begin{subfigure}[b]{0.45\textwidth}
\includegraphics[width=\textwidth]{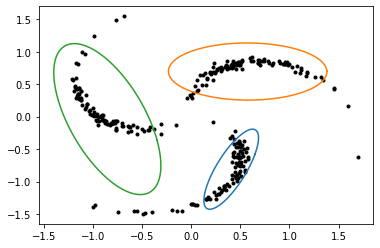}
        \caption{Set 4 - (0.886,800.70)}
    \end{subfigure}

 \begin{subfigure}[b]{0.45\textwidth}
\includegraphics[width=\textwidth]{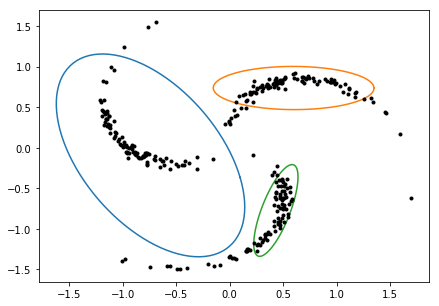}
        \caption{Set 4 - (0.781,805.28)}  
\end{subfigure}
   ~ 
\begin{subfigure}[b]{0.45\textwidth}
\includegraphics[width=\textwidth]{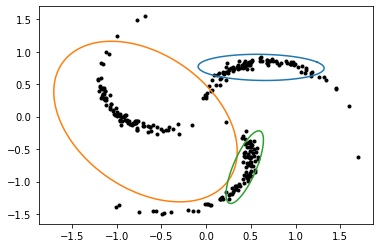}
        \caption{Set 4 - (0.722,806.93)}
    \end{subfigure}
       \begin{subfigure}[b]{0.45\textwidth}
\includegraphics[width= \textwidth]{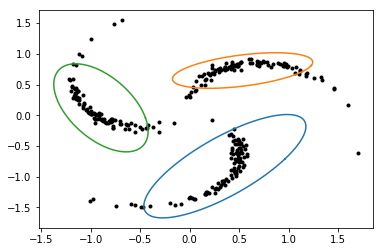}
        \caption{Set 4 - (\textbf{0.941},813.40)}
    \end{subfigure}%
    ~
     \begin{subfigure}[b]{0.45\textwidth}
\includegraphics[width= \textwidth]{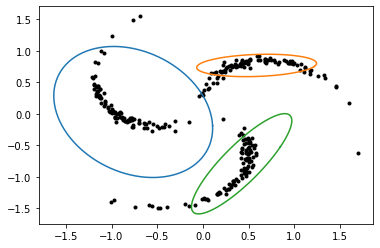}
        \caption{Set 4 - (0.778,846.99)}
    \end{subfigure}%
	\caption{8 clustering solutions obtained with AD-GD using different initializations on Pinwheel data (with parameters $r=0.3, t=0.05, s=0.4$); Sets refer to groups defined in Table \ref{tab:missstats}; in parentheses: (ARI, AIC),  best values in bold.}
	\label{fig:SOGD_manifold}      
    \end{minipage}
\end{figure}

We study the clustering solutions obtained by fitting misspecified GMM on 
Pinwheel data, also called warped GMM, with 3 components and 2 dimensions
\citep{iwata2013warped}.
Pinwheel data is generated by sampling from Gaussian distributions, and then stretching and rotating the data. %
The centers are equidistant around the unit circle. 
The variance is controlled by two parameters $r$ and $t$, the radial standard deviation and the tangential standard deviation respectively. 
The warping is controlled by a third parameter, $s$, the rate parameter. 
Thus, the extent of misspecification (i.e., the deviation of the data from the assumed Gaussian distributions in GMM-based clustering) can be controlled using these parameters.
An example is shown in fig. \ref{fig:SOGD_manifold}.
We generate 1800 Pinwheel datasets with different combinations of parameters.
In addition, we also simulate 1800
3-component, 2-dimensional datasets from GMM with varying overlap of components (to analyze the case where there is no misspecification in fitting a GMM).
Details of both Pinwheel and GMM datasets used are given in Appendix \ref{app:robust_initialization}. 
For each dataset we obtain two clustering solutions by using EM and AD-GD.
We run both the algorithms using the same initialization and stopping criterion (convergence threshold $1e-5$). 
We use %
ARI \citep{hubert1985comparing}
to evaluate performance where higher values indicate better clustering. 

\begin{figure}[!h]
    \centering
    \begin{minipage}{.5\textwidth}
        \centering
    \begin{subfigure}[b]{0.5\textwidth}
\includegraphics[width= \textwidth]{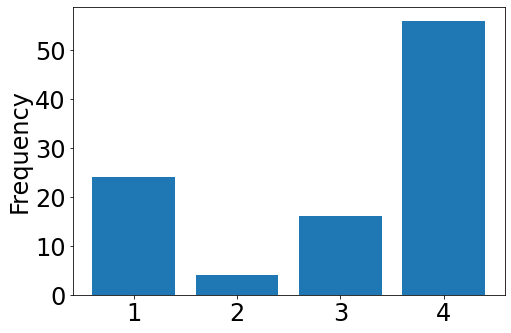}
        \caption{EM}    
    \end{subfigure}%
    \begin{subfigure}[b]{0.5\textwidth}
\includegraphics[width=\textwidth]{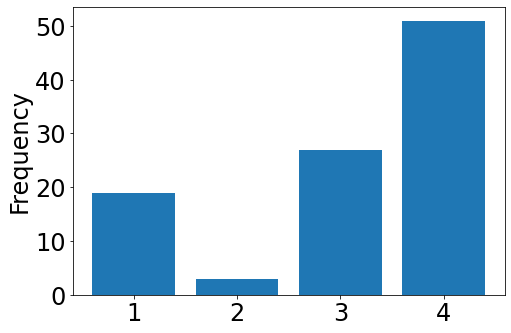}
        \caption{AD-GD}  
    \end{subfigure}
    \caption{Frequency over 100 random initializations}
    \label{fig:frequencies}
    \end{minipage}
    \end{figure}

Our extensive experiments on these 3600 datasets, described in Appendix \ref{app:robust_initialization}, show that EM outperforms AD-GD in both cases -- when there is misspecification and no missspecification.
However, when there is misspecification, we find that in both EM ad AD-GD, there are many {\it inferior} solutions
that have low ARI and unequal fitted covariances, often with one fitted component having relatively large covariance, resulting in a high degree of overlap between components.
We illustrate this using Pinwheel data generated with parameters $r=0.3$, $t=0.05$ and $s=0.4$.
Both EM and AD-GD are run with 100 different initializations.
Table \ref{tab:missstats} shows the summary statistics of the AD-GD solutions where we have grouped the solutions into 4 sets.
Appendix \ref{appendix:binning} has statistics of the EM solutions that are very similar.
We categorize all the solutions with AIC higher than 788 into the fourth 
set.
Fig. \ref{fig:frequencies} shows the histogram of the counts in the 4 sets.
We visualize the clustering for a single solution from sets 1--3 and 5 solutions from set 4 in fig. \ref{fig:SOGD_manifold}.

\begin{table*}[!h]
\centering
\captionof{table}{Summary statistics of AD-GD clustering solutions over 100 random initializations on the Pinwheel dataset (shown in fig. \ref{fig:SOGD_manifold}), grouped into 4 sets based on AIC ranges. Mean and standard deviation of AIC, ARI, component weights and covariance determinants, over solutions in each set.}
\label{tab:missstats}
\begin{tabular}{cc|p{1.35cm} p{1.25cm}p{1.25cm} p{1.25cm} p{1.25cm}p{1.35cm} p{1.25cm} p{1.25cm} } 
 \hline
Set & AIC Range & AIC & ARI & $\pi_1$ & $\pi_2$ & $\pi_3$  & $|\bSigma_1|$ & $|\bSigma_2|$ & $|\bSigma_3|$  \\ \hline %
 
1 & 771-773 & 771.9 \newline (6e-8) & 0.625 \newline (2e-16) & 0.257 \newline(3e-6) & 0.265 \newline(3e-6) & 0.477 \newline(1e-6) & 0.0002 \newline (1e-9) & 0.0005 \newline (4e-9) & 0.123 \newline (4e-7) \\ %
2 &  781-782 & 781.1 \newline (3e-6) & 0.912 \newline (0) & 0.306 \newline (1e-5) & 0.341 \newline (3e-5) & 0.352 \newline (1e-5) & 7e-4 \newline(5e-9) & 0.01 \newline(5e-9) & 0.01 \newline(5e-7) \\ %
3 &  786-787 & 786.8 \newline(2e-7) & 0.652 \newline (0) & 0.257 \newline (5e-6) & 0.267 \newline (3e-6) & 0.475 \newline (3e-6) & 2e-4 \newline 3e-9 & 5e-4 \newline (3e-9)  & 0.156 \newline (2e-6) \\ %
4 &  788-850 & 815.0 \newline(17.84)  & 0.806 \newline(0.06) & 0.28 \newline (0.015) & 0.315 \newline (0.010) & 0.403  \newline (0.024) & 6e-4 \newline (4e-4) & 3e-3 \newline(1e-3)  & 0.047 \newline(0.018)  \\ \hline 
\end{tabular}
\end{table*}

We observe that both EM and AD-GD obtain very similar solutions in terms of AIC and ARI for this dataset.
The best average ARI 
is that of set 2 (row 2 of Table \ref{tab:missstats}).
Fig. \ref{fig:frequencies} shows that these solutions 
are obtained in less than 5\% of the cases.
The overall best ARI, of 0.941, is from a solution in set 4 that has a high AIC value of 813.4 as shown in fig. \ref{fig:SOGD_manifold} (g).
The best AIC, of 771.9, is obtained by a solution from set 1 which has considerably lower ARI of 0.625 as shown in fig. \ref{fig:SOGD_manifold} (a).
Thus, high likelihood does not correspond to a good clustering.
We observe that there are many inferior solutions in sets 1, 3 and 4 having a fitted component with large variance and low ARI, also seen in the specific solutions shown in fig. \ref{fig:SOGD_manifold}.

We find that such inferior solutions with low ARI and low AIC in misspecified models occur frequently with many different initializations,
and typically when there is a component with large variance. 
This is different from the characterization of spurious solutions that were found to occur rarely, only with certain initializations and due to a component with small variance
\citep{McLa:Peel:fini:2000}.

\subsection{High Dimensional Data}
\label{section:dominating}
EM-based methods fail to run in high-dimensional low sample size regime. 
We observe that the clustering solutions from AD-GD 
often have a single component with relatively low variance and high mixture weight spread over all the datapoints.
We call such components `dominating' components.
While the %
final dominating component depends on the initialization,
one such component is present for almost all the initializations. 
Note that while the characterization of this clustering is also different from that of previously studied spurious solutions, the effect remains the same -- that of an inferior clustering.
Here the likelihood increases predominately by increasing the mixture component weight of the dominating component.

We illustrate the dominating components through the determinants of covariance matrices, using a simulated 200-dimensional 4-component dataset.  Each component has covariance matrix  $\bSigma_k = 0.5*\mathbb{I}$. The mean of the components are chosen as shown in table \ref{table:dom_illus_mean}.
For each component, we sample 15 datapoints. 
\begin{table}[h]
\centering
\captionof{table}{Component (C) means}
\label{table:dom_illus_mean}
\begin{tabular}{ccccc} 

 \hline
 Features &  \text{C-1} & \text{C-2}  & \text{C-3} & \text{C-4} \\ \hline
1-20 & 1 & 0  & 0 & 0 \\ 
20-40 & 0 & 1 &  0  & 0 \\ 
40-60 & 0  & 0 & 1 & 0 \\
60-80 & 0 & 0 & 0 & 1
\\
80-200 & 0 & 0 & 0 & 0 \\

 \hline
 
\end{tabular}
\end{table}

\begin{figure}[h]
\centering
\includegraphics[width= 0.4\textwidth]{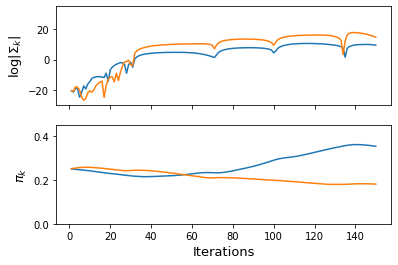}

	\caption{In HD ($p > n$): (a) maximizing likelihood
	leads to reduction in variance (above) and corresponding increase in weight (below) resulting in a \textit{dominating component} (in blue). GMM (AD): ARI = 0.037 Loglikelihood = -24416.}
	\label{fig:ad-gd_hd}
\end{figure}

Fig. \ref{fig:ad-gd_hd} illustrates the dominating component phenomenon over multiple iterations: soon after initialization, the variance of one of the components (the dominating component) increases to capture all the near-by points and then, the variance of this component is reduced while at the same time the corresponding component weight is increased, to maximize the likelihood.

\begin{figure*}[!htb]
\centering     
 \begin{subfigure}[b]{0.19\textwidth}
\includegraphics[width=\textwidth]{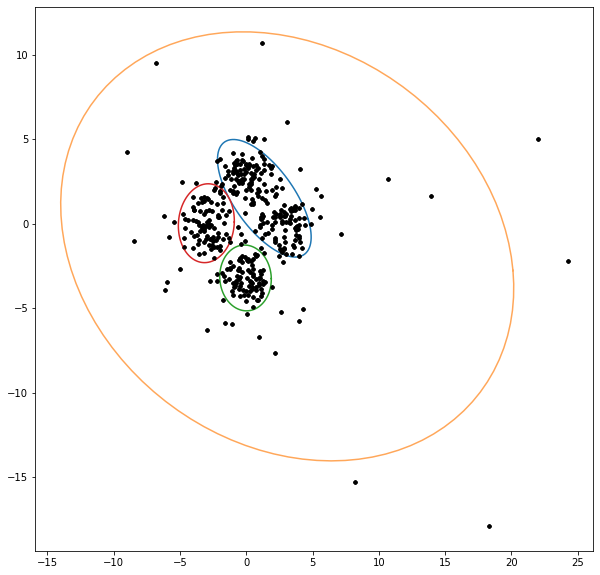} 
        \caption{D.o.F = 2} 
    \end{subfigure}
 \begin{subfigure}[b]{0.19\textwidth}
\includegraphics[width=\textwidth]{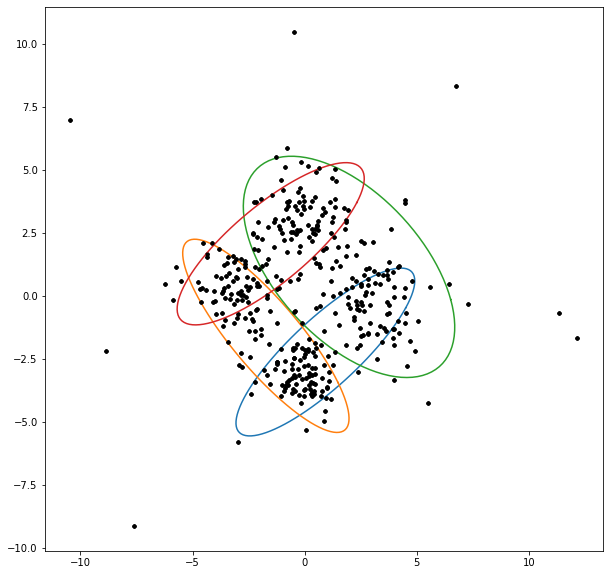} 
        \caption{D.o.F = 3}
    \end{subfigure}
     \begin{subfigure}[b]{0.19\textwidth}
\includegraphics[width=\textwidth]{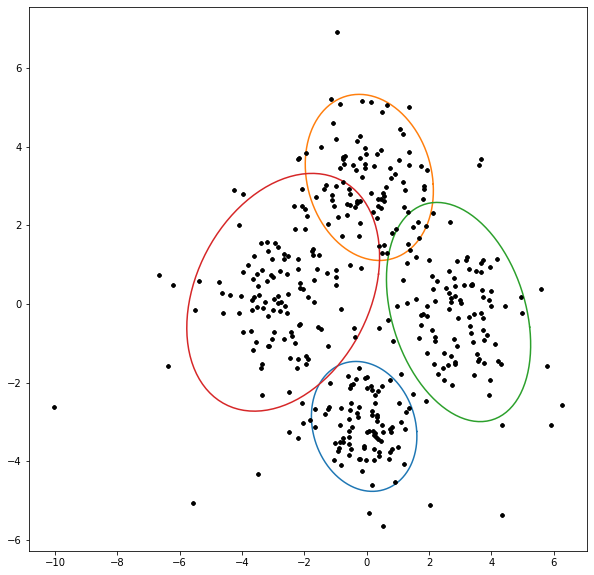} 
        \caption{D.o.F = 4}
    \end{subfigure}
 \begin{subfigure}[b]{0.19\textwidth}
\includegraphics[width=\textwidth]{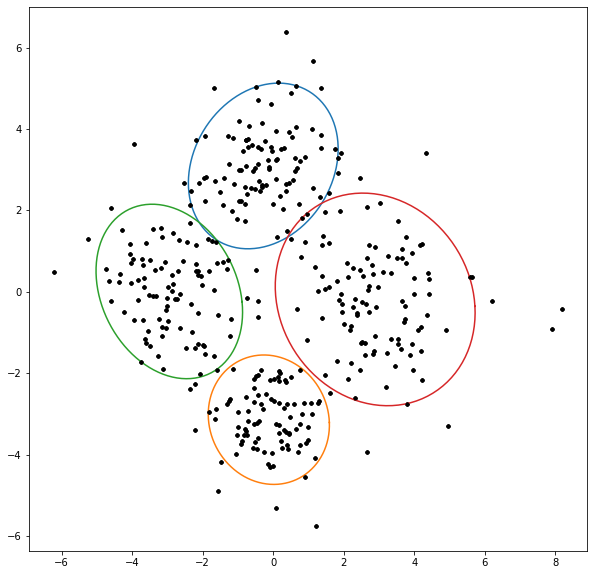} 
        \caption{D.o.F = 5}
    \end{subfigure}
     \begin{subfigure}[b]{0.19\textwidth}
\includegraphics[width=\textwidth]{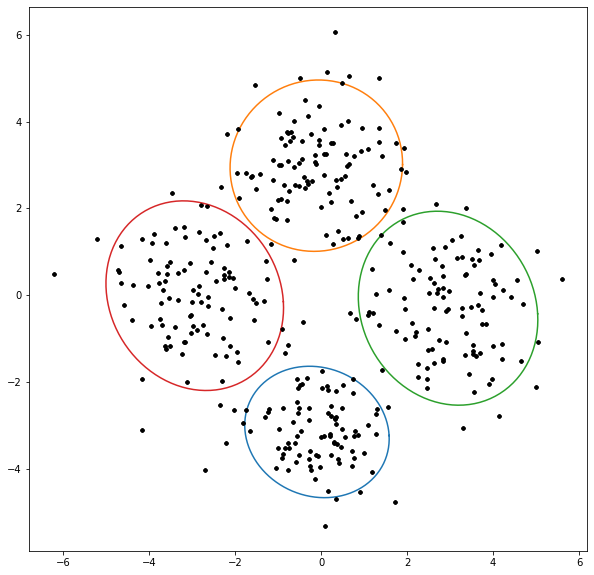} 
        \caption{D.o.F = 10}
    \end{subfigure}
	\caption{ Lower the degree of freedom (D.o.F), higher the misspecification, greater the difference in orientation and size of the components fitted}
	\label{fig:corruption}
\end{figure*}
\begin{figure*}[!h]
\centering
 \begin{subfigure}[b]{0.19\textwidth}
\includegraphics[width=\textwidth]{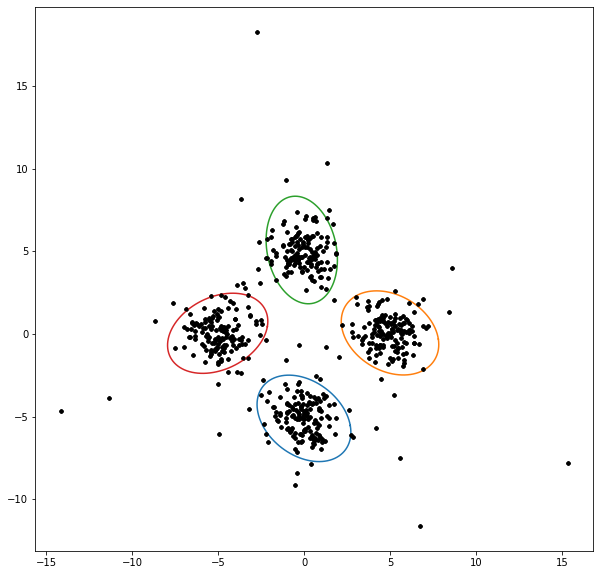} 
        \caption{D.o.F = 2}
        
    \end{subfigure}
         ~
 \begin{subfigure}[b]{0.19\textwidth}
\includegraphics[width=\textwidth]{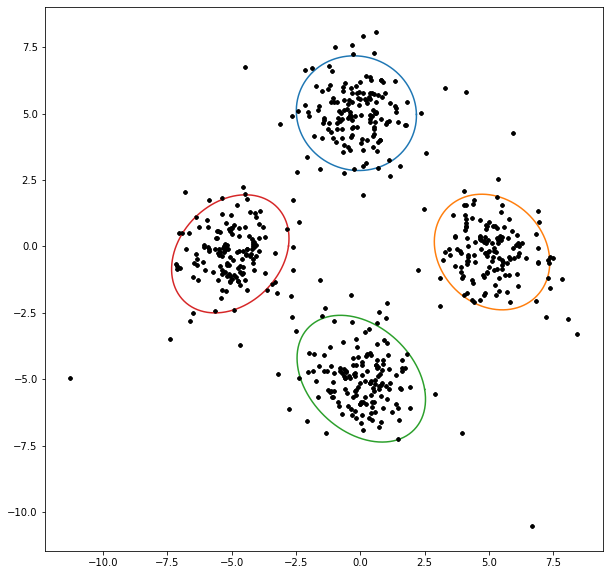} 
        \caption{D.o.F = 3}
        
    \end{subfigure}
    ~
     \begin{subfigure}[b]{0.19\textwidth}
\includegraphics[width=\textwidth]{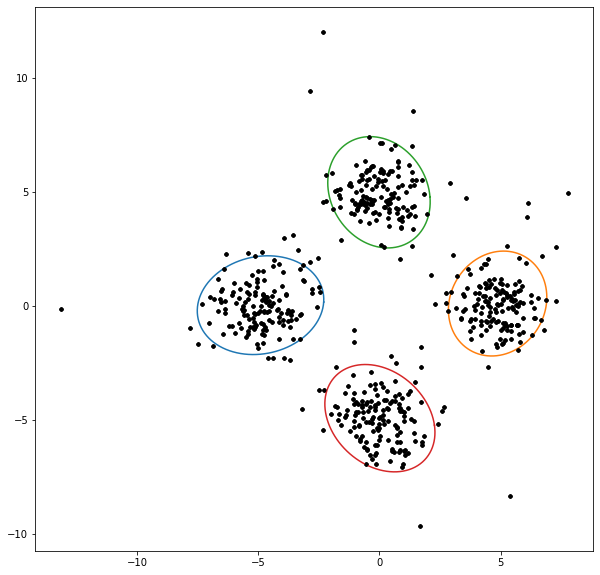} 
        \caption{D.o.F = 4}
        
    \end{subfigure}
      ~
 \begin{subfigure}[b]{0.19\textwidth}
\includegraphics[width=\textwidth]{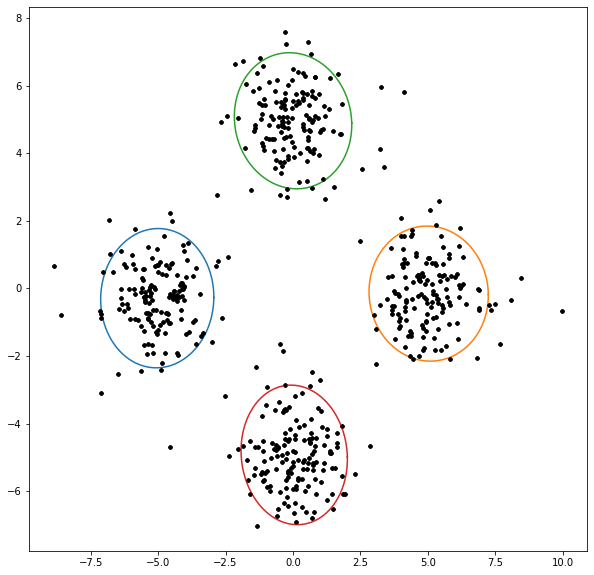} 
        \caption{D.o.F = 5}
        
    \end{subfigure}
    ~
     \begin{subfigure}[b]{0.19\textwidth}
\includegraphics[width=\textwidth]{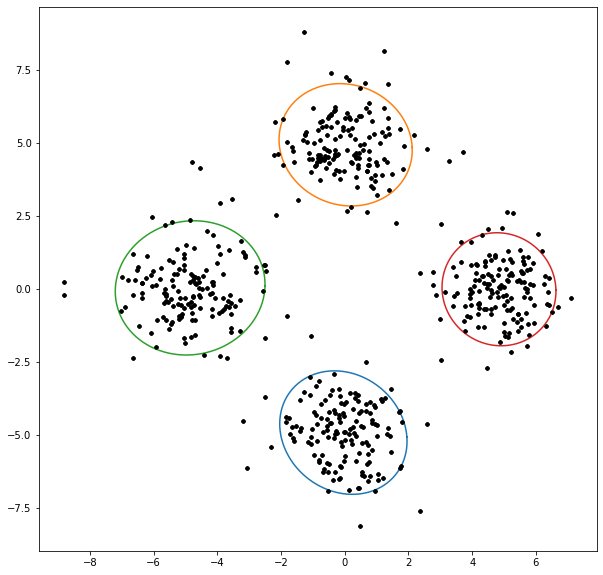} 
        \caption{D.o.F = 10}
    \end{subfigure}
    
	\caption{ When the cluster separation is high, the misspecification does not seem to affect the fitted components much - only the orientation of the fitted components is affected}
	\label{fig:corruption_t_well_separated}
\end{figure*}

\section{A Penalty Term based on KL Divergence}
\label{sec:penalty}

Our empirical analysis in the previous section reveals that  clustering solutions from AD-GD tend to 
have inferior solutions in cases of misspecification and high dimensional data. 
Although their characteristics are different from previously studied spurious solutions, they are similar in terms of being poor clustering solutions that have high likelihood.
A well-known technique to avoid spurious solutions is to constrain the parameter space
\citep{Hath:Acon:1985}).
Approaches include restrictions on scale and weight parameters, e.g., through bounds on the minimum eigenvalues of component covariance matrices \citep{Ingr:Rocc:2007,ingrassia2011degeneracy} or constraining their determinants 
\citep{biernacki2004asymptotic} during EM estimation.
An approach based on penalized ML estimators was proposed by 
\cite{Chen2009}.
They design a family of penalty terms that ensures consistency of the estimator.
An example of a penalty term that belongs to their family is:
$\sum_{k=1}^K -\frac1n\left(\operatorname{tr}\left(\bSigma_k^{-1}\right)+\log |\bSigma_k|\right)$.
We design a penalty term for the likelihood that is similar in spirit, but more general, to address the problem of inferior clustering solutions.

\subsection{Motivating Examples}

In cases of misspecification, inferior clusterings from both EM and AD-GD, are found to 
have fitted covariances that differ considerably in their orientations and sizes.
We observe this in the solutions in 
fig. \ref{fig:SOGD_manifold}

As another motivating example, consider a case where misspecification arises due to contamination \citep{punzo2016parsimonious}.
Consider a 2-dimensional GMM with 4 components, each with unit covariance. 
The means of the four components are set as given in table   \ref{tab:t} with $\beta =3$.
\begin{table}[!h]
\caption{Component (C) Means}
\centering
\begin{tabular}{ccccc}
\hline
& C-1 & C-2 & C-3 & C-4 \\ \hline
$\mu_1$ & -$\beta$ & 0 & 0 & $\beta$ \\
$\mu_2$ & 0 & $\beta$ & -$\beta$ & 0 \\ \hline
\end{tabular}
\label{tab:t}
\end{table}
From each component, we sample 50 datapoints. 
Now, we consider a 2-dimensional 4-component multivariate $t$ distribution with same means and covariance structures. 
For the mixture of multivariate $t$ distribution also, we sample 50 datapoints from each component. 
Each dataset has 400 samples. 
We thus have a contaminated Gaussian mixture.
When the degree of freedom is high, the contamination is negligible. When the degree of freedom is low, the $t$-distributions components will have heavier tails, hence the contamination is higher.  
By varying the degrees of freedom we can control the contamination and hence the misspecification.  
We fit a 4-component GMM using EM (We obtain similar results using AD-GD as well). As seen from figure \ref{fig:corruption}, the components tend to overlap with increasing misspecification (decreasing D.o.F). We observe that when the misspecification is high, some of the fitted components tend to have higher covariance compared to other components.

We repeat the same experiment of corrupting the GMM with a multivariate $t$-distribution but with increasing the cluster separation by setting $\beta=5$.
The results are shown in fig. \ref{fig:corruption_t_well_separated}. As seen, increasing cluster separation mitigates the affect of misspecification. A similar experiment on contamination with random noise is shown in Appendix \ref{section:appendix_noise}.

\subsection{Combinatorial KL Divergence}
We propose a novel penalty term that penalizes differences in orientations and sizes of the fitted components as well as degenerate solutions.
Our penalty term is based on the KL-divergence between component Gaussians.
Let $\mathcal{N}_k$ denote the multivariate Gaussian distribution $\mathcal{N}(\bmu_k,\bSigma_k)$.
The KL-divergence between $\mathcal{N}_1$ and $\mathcal{N}_2$
is given by: 
 \begin{dmath*}
    KL\left(\mathcal{N}_1, \mathcal{N}_2\right) = 
    KL\left(\mathcal N(\bmu_1,\bSigma_1),\mathcal N(\bmu_2,\bSigma_2) \right) = \frac{1}{2}\left[\log\frac{|\bSigma_2|}{|\bSigma_1|} - p + \text{tr} \{ \bSigma_2^{-1}\bSigma_1 \} + (\bmu_2 - \bmu_1)^T \bSigma_2^{-1}(\bmu_2 - \bmu_1)\right]
\end{dmath*}
Each term in the expression above can provide useful constraints as summarized in 
Table \ref{tab:kl_intuitioin}.
\begin{table}[!h]
\captionof{table}{Purpose served by each term in $KL\left(\mathcal{N}_1, \mathcal{N}_2\right)$}
	\centering
\begin{tabular}{cp{4cm}} 

 \hline
 Penalty term &  Purpose  \\ \hline
 $\tr{(\bSigma_1^{-1}\bSigma_2)} - p$ & This term penalizes the difference in orientations, i.e. if the directions of principal axes of the covariance matrices of the two components are vastly different. When $\bSigma_1$ and $\bSigma_2$ are exactly similar, this penalty term goes to zero. \\
 \hline
 $\log\frac{|\bSigma_2|}{|\bSigma_1|}$ & This term penalizes the difference in size of the covariance matrices of the two components. Note that this term can go to zero even if the orientations of two covariance matrices are different.   \\
 \hline
  $(\bmu_2 - \bmu_1)^T \bSigma_2^{-1}(\bmu_2 - \bmu_1)$ & This term penalizes the assignment of a single cluster to faraway outlier points \\ \hline

 \hline
 
\end{tabular}
\label{tab:kl_intuitioin}
\end{table}

KL-divergence is not symmetric about $\bSigma_1$ and $\bSigma_2$.
If there is an order of magnitude difference in the covariance matrices $\bSigma_1$ and $\bSigma_2$, it can be detected through the values of
$KL\left(\mathcal{N}_1, \mathcal{N}_2\right)$
and $KL\left(\mathcal{N}_2, \mathcal{N}_1\right)$.
The difference in their values is primarily contributed by the difference between the  terms $(\bmu_2 - \bmu_1)^T \bSigma_2^{-1}(\bmu_2 - \bmu_1)$ and  $(\bmu_2 - \bmu_1)^T \bSigma_1^{-1}(\bmu_2 - \bmu_1)$, and $ \text{tr} \{ \bSigma_2^{-1}\bSigma_1 \} $ and $ \text{tr} \{ \bSigma_1^{-1}\bSigma_2 \}$.
The difference in these two KL divergence values provides signal about the overlap or asymmetry of the two Gaussians.
This notion is generalized to a $K$-component GMM, through the combinatorial KL divergences, KLF (forward) and KLB (backward): 
\begin{align}
 KLF  := \sum_{1 \le k_1 < k_2 \le K} KL\left(\mathcal{N}_{k_1}, \mathcal{N}_{k_2}\right);\, \\
KLB  := \sum_{1 \le k_2 < k_1 \le K} KL\left(\mathcal{N}_{k_1}, \mathcal{N}_{k_2}\right).
\end{align}

Well separated clusters typically have equal and lower values of KLF and KLB. 
We denote both the values by $$KLDivs = \{KLF,KLB\}.$$
We note that these two sums (KLF + KLB) together give the sum of Jeffrey's divergence between all components. 
In the clustering outputs shown in fig. \ref{fig:SOGD_manifold}, 
we see that in solution (c) from set 3, where clustering is poor, KLF$=258$ and KLB$=494$ (the difference is high), while in solution (g) from set 4, which has better clustering, KLF$=127$ and KLB$=64$ (with relatively lower difference).

Our proposed penalty term is a weighted sum of the $KLF$ and $KLB$ terms:
\begin{equation}
- w_1 \times KLF - w_2 \times KLB
\end{equation}
Notice that the weights $(-w_1,-w_2)$ are negative.
If we add positive weights GD will further shrink the smaller clusters. 
Negative weights lead to separation as well as clusters of similar volume.
In high dimensions, the variance of the non-dominating components increase to reduce KLF and KLB
(for an example, see section \ref{sec:illus}).
To tackle this problem, we add another term
to penalize very small covariance matrices: %
$-w_3 \times \sum_{k = 1}^K (\det(\bSigma_k) - \lambda_k)^2$.

\subsection{Implementation}

Optimization of likelihood with these penalty terms can be implemented effortlessly in the AD-GD framework where gradients in closed forms are not required.
However, the use of such complex penalties is difficult within EM.
We cannot obtain closed forms of the covariance estimates.
Closed forms for the mean update can be derived (Appendix \ref{appendix:closed_forms}) but is laborious.
Further the update for the mean of a component depends on means for all other components, and hence cannot be parallelized in EM.

\section{Sequential Initialization Algorithm}
\label{sec:SIA}

Our algorithm is similar in principle to that of \cite{biernacki2003choosing} who suggest the use of initializing EM with a few short runs of EM.
In our algorithms, after the first run that can use EM or AD-GD without the penalty term, we run AD-GD with the penalty term.
Note that for high-dimensional data, the first run requires AD-GD.

\subsection{SIA}
SIA consists of two steps.
In the first step of SIA, we use the loglikelihood $\mathcal{L}$ as the objective and run EM or AD-GD (first or second order) to fit a GMM.
Typically, the output at the end of first step will have unequal KL-divergences for misspecified models. 
We take these parameters at the end of the first step to initialize the algorithm in the second step. 
In the second step we modify the objective function to:
\begin{equation}
\label{eq:SIA}
\mathcal{M} =  \mathcal{L} - w_1 \times KLF - w_2 \times KLB   
\end{equation}
After the second optimization step the likelihood decreases but the KL-divergence values, KLF and KLB, come closer. 
Thus there is a trade off between likelihood and cluster separation.
In high-dimensions, the objective function $\mathcal{M}$ is set to:
\begin{equation}
\label{eq:SIA-HD}
\mathcal{M} =  \mathcal{L}  - w_1 \times KLF -w_2 \times KLB - w_3\times \sum_{k = 1}^K (\det(\bSigma_k) - \lambda_k)^2
\end{equation}

To distinguish between the low-dimensional and high-dimensional cases, we call the latter SIA-HD. 
The complete algorithm is presented in Algorithm \ref{ALGO:SIA_combined}.

\begin{algorithm}[t!]
\caption{SIA}
\label{ALGO:SIA_combined}

\textbf{Input: }
Data: $n \times p$ dimensional matrix, 
required number of clusters $K$, 
learning rate $\epsilon$, 
convergence tolerance $\gamma$.

\textbf{Initialize at iteration $t=0$:} 
$\hat{\bmu}^{0}_k, \hat{\alpha}_k$'s using K-Means or random initialization; 
$\hat{\bU}^{0}_k$ (for GMM) or $\hat{\bLambda}^0_k, \hat{\bPsi}^0_k$ (for MFA) as identity matrices.

\textbf{Step I:} Run Algorithm \ref{ALGO:AD-GD} (AD-GD)

\textbf{Step II:} 

\textbf{For Low Dimensional Data:}  Set\\ $ \mathcal{M} =  \mathcal{L} - w_1 \times KLF -w_2 \times KLB$

\textbf{For High-Dimensional Data: } 
Use the determinant of $\bSigma_k$'s to identify dominating components and set penalty terms $\lambda_k$. Set

\resizebox{.9\linewidth}{!}{
  \begin{minipage}{\linewidth}
\begin{align*}
\mathcal{M} =   \mathcal{L} - w_1 \times KLF -w_2 \times KLB   - w_3 \times \sum_{k = 1}^K (\det(\bSigma_k) - \lambda_k)^2
\end{align*}
\end{minipage}
}
\textbf{Initialize:} Use Output of Step I

\textbf{REPEAT:} At every iteration $t+1$:

\resizebox{.9\linewidth}{!}{
  \begin{minipage}{\linewidth}
  \begin{align*}
\hat{\alpha}^{t+1}_k := \hat{\alpha}^{t}_k + \epsilon \frac{\partial \mathcal{M} }{\partial \alpha_k} ; \; \hat{\pi}^{t+1}_k  := \frac{ e^{\hat{\alpha}^{t+1}_k} }{\sum_{k^'} e^{\hat{\alpha}^{t+1}_{k^'}}} ; \;  \hat{\bmu}^{t+1}_k := \hat{\bmu}^{t+1}_k + \epsilon  \frac{\partial \mathcal{M}}{\partial \bmu_k} ; \;
\end{align*}
  \end{minipage}
}

\resizebox{\linewidth}{!}{
  \begin{minipage}{\linewidth}
  
\textbf{Covariance updates for GMM:}
  \begin{align*}
    \hat{\bU}^{t+1}_k := \hat{\bU}^{t}_k + \epsilon \frac{\partial  \mathcal{M}}{\partial \bU_k} ; \;
    \hat{\bSigma}^{t+1}_k := \hat{\bU}^{t+1}_k \hat{\bU}^{{t+1}^T}_k 
    \end{align*}
 
\textbf{Covariance updates for MFA:}   
    \begin{align*}
    \hat{\bLambda}^{t+1}_k  := \hat{\bLambda}^{t}_k + \epsilon \frac{\partial \mathcal{M}}{\partial \bLambda_k}; 
    \hat{\bPsi}^{t+1}_k := \hat{\bPsi}^{t}_k + \epsilon  \frac{\partial \mathcal{M}}{\partial \bPsi_k}; 
    \hat{\bSigma}^{t+1}_k := \hat{\bLambda}^{t+1}_k \hat{\bLambda}^{{t+1}^{T}}_k + \hat{\bPsi}^{t+1}_k
\end{align*}
  \end{minipage}
}

 \textbf{UNTIL:} convergence criterion $|\mathcal{M}^{t+1} - \mathcal{M}^{t} | < \gamma $ is met

\end{algorithm}

\subsection{MPKL: A Model Selection Criterion}
We define the maximum absolute pairwise difference between KL divergence values (MPKL) for a $K$-component GMM as follows:
\begin{align}
\label{eq:MPKL}
MPKL = \max_{1 \le k_1,k_2 \leq K} | KL\left(\mathcal{N}_{k_1}, \mathcal{N}_{k_2}\right) - KL\left(\mathcal{N}_{k_2}, \mathcal{N}_{k_1}\right)|
\end{align}
As discussed earlier, it is an indicator of how well the separation between clusters is.
This can be used as a criterion for selecting the number of clusters.
For a chosen range of number of clusters ($2,\ldots,L$), we compute MPKL for each value and choose $K$ that minimizes MPKL:
$$\text{argmin}_{K \in [2,\ldots,L]} MPKL$$

\subsection{Computational Complexity}
The computational complexity is dominated by evaluating $KLF$ and $KLB$ which involves $O(K^2)$ matrix inversion steps ($O(p^3)$). Therefore, the overall computational complexity of all our algorithms (SIA,SIA-HD and model selection) is $O(K^2p^3)$.  
Appendix \ref{app:complexity} presents a comparison of runtime with EM and AD-GD. 

\subsection{An Upper Bound on the SIA Objective Function}
\label{section:theoretical_analysis}
Consider fitting a two-component GMM and without loss of generality, assume, $|\hat \bSigma_1^{t}| \leq |\hat \bSigma_2^{t}|$, at some iteration $t$. 
We also assume that the spectral norm of $\hat \bSigma_1^{t}, \hat \bSigma_2^{t}$ is bounded by some $c < \infty$ as a regularity condition. 
Let $C$ denote a constant dependent only on $p,n,w_1,w_2$ and $c$ and independent of $\btheta$.
Let $\mathcal{M}$ denote the penalized log-likelihood (this result holds for both SIA and SIA-HD).
We prove that 
the use of the penalty terms leads to a bounded objective function.
\begin{theorem}
For non-negative weights $w_1,w_2$
, and any iteration $t$, 
$\mathcal{M} \le
- \frac{n}{2} (1+w_2)\left(  \log (\frac{w_2}{1+w_2} \hat \bSigma_2^{t}) + p  \right) + C.$
\end{theorem}

The proof is in Appendix \ref{app:proof_lemma1}. %
Also note that the bound is given in terms of $w_2$ because we assume $|\hat \bSigma_1^t| \leq |\hat \bSigma_2^t|$. 
By setting $w_2 = 0$, we can see that the maximum of unpenalized likelihood is unbounded.
If a $\bSigma$ collapses, %
the log-likelihood may increase
but
the trace term and the determinant in the penalization will explode. Hence, by having negative weights $-w_1, -w_2$ %
we can ensure that $\mathcal{M} $ %
is bounded and control the behavior of the algorithm. 
A visualization of the likelihood surface, with and without penalty terms, is shown in Appendix \ref{sec:visu}.

\subsection{Illustrative Examples}
\label{sec:illus}

\begin{figure}[h!]
\centering
\captionsetup{justification=centering}
    \begin{subfigure}[b]{0.25\textwidth}
\includegraphics[width= \textwidth]{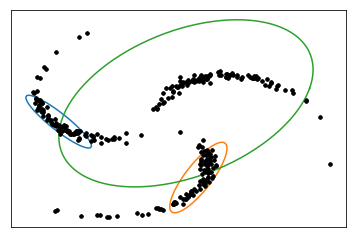}
        \caption{After Step 1:\\ ARI = 0.65\\ KLDivs = \{258,494\},\\ Loglikelihood = -376}
        
    \end{subfigure}%
    ~
 \begin{subfigure}[b]{0.25\textwidth}
\includegraphics[width=\textwidth]{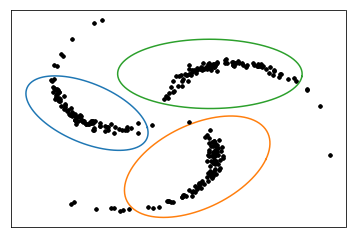}
        \caption{After Step 2:\\ ARI = 0.96 \\  KLDivs  = \{37,36\}, \\ Loglikelihood = -437 }
        
    \end{subfigure}
	\caption{Clustering using SIA in 3-component pinwheel data. Compare with figure \ref{fig:SOGD_manifold}.}
	\label{fig:sia_4comp}
\end{figure}

We show how SIA improves the clustering performance in the three specific examples discussed in earlier sections.
Fig. %
\ref{fig:sia_4comp} shows the clustering obtained on the %
Pinwheel data discussed in section \ref{sec:autodiff}.
We observe the difference in the clustering after the first and second steps of SIA.
After the second step, compared to the clustering after step 1, the likelihood decreases, both the KLF and KLB values decrease, the ARI increases, and the clusters have less overlap.

\begin{figure}[!h]
\centering
\captionsetup{justification=centering}

 \begin{subfigure}[b]{0.5\textwidth}
\includegraphics[width=\linewidth]{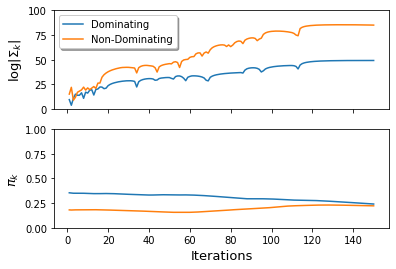}
        \caption{ SIA: ARI = 0.048 \\   Loglikelihood = -19023 }
    \end{subfigure}%
    
    \begin{subfigure}[b]{0.5\textwidth}
\includegraphics[width= \linewidth]{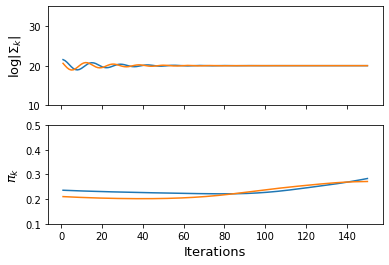}
        \caption{SIA-HD: ARI = 0.35 \\ Loglikelihood = -14807 }
    \end{subfigure}
	\caption{ In HD ($p >> n$): (a) SIA: KLF and KLB  are reduced by increasing the variance of non-dominating component in SIA, (b) SIA-HD: there is no dominating component in the final clustering.}
	\label{fig:sia2_hd}
\end{figure}

\begin{figure*}[!htb]
\centering
 \begin{subfigure}[b]{0.19\textwidth}
\includegraphics[width=\textwidth]{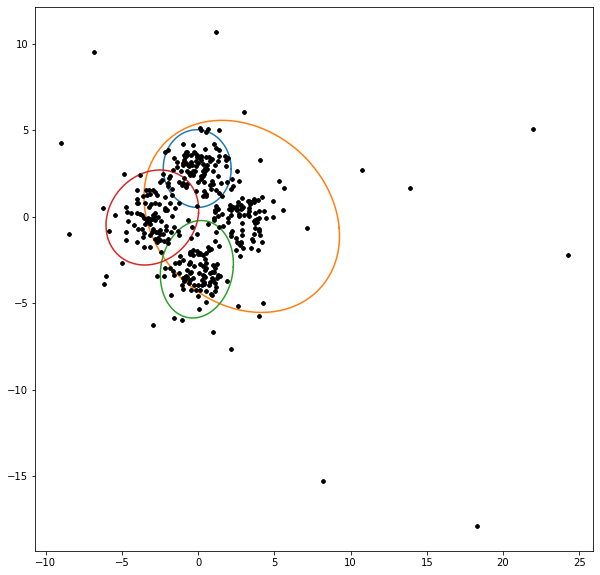} 
        \caption{D.o.F = 2}
        
    \end{subfigure}
         ~
 \begin{subfigure}[b]{0.19\textwidth}
\includegraphics[width=\textwidth]{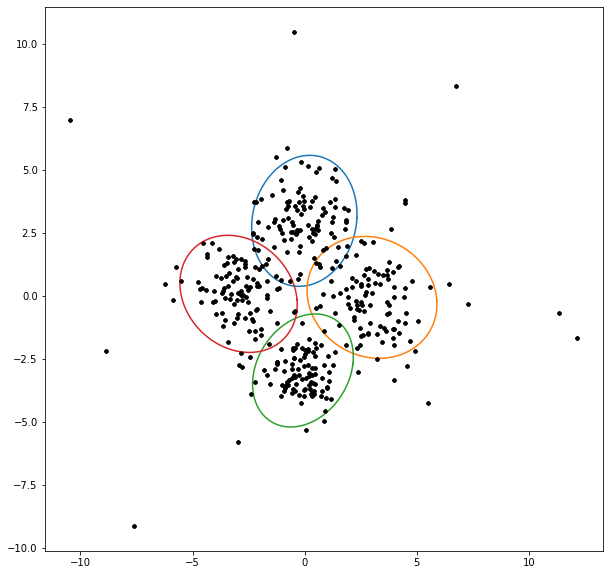} 
        \caption{D.o.F = 3}
        
    \end{subfigure}
    ~
     \begin{subfigure}[b]{0.19\textwidth}
\includegraphics[width=\textwidth]{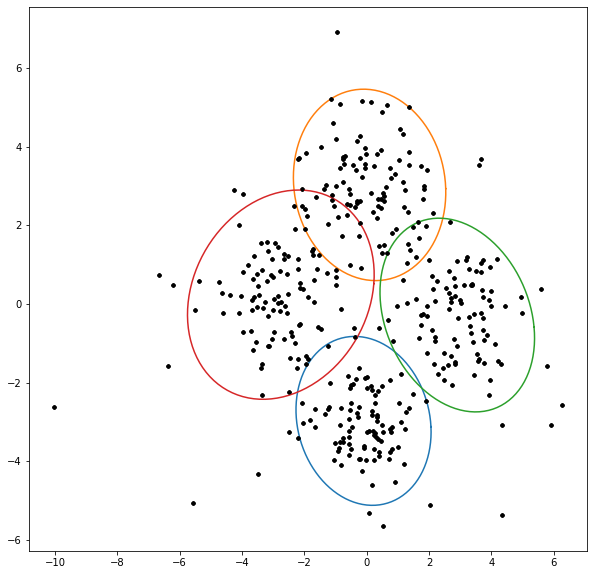} 
        \caption{D.o.F = 4}
        
    \end{subfigure}
      ~
 \begin{subfigure}[b]{0.19\textwidth}
\includegraphics[width=\textwidth]{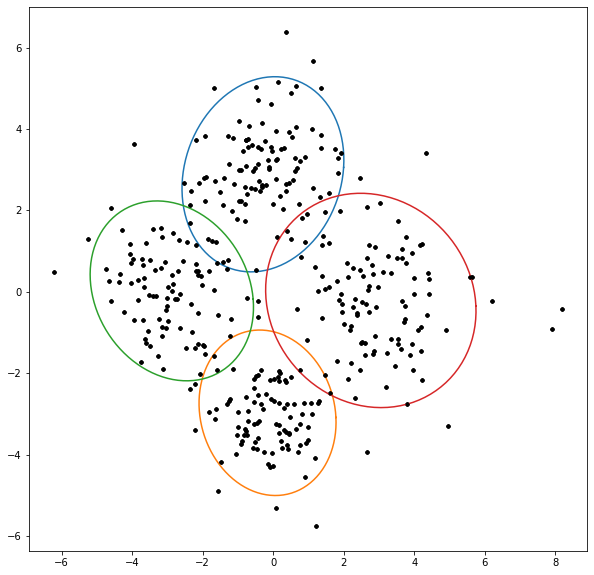} 
        \caption{D.o.F = 5}
        
    \end{subfigure}
    ~
     \begin{subfigure}[b]{0.19\textwidth}
\includegraphics[width=\textwidth]{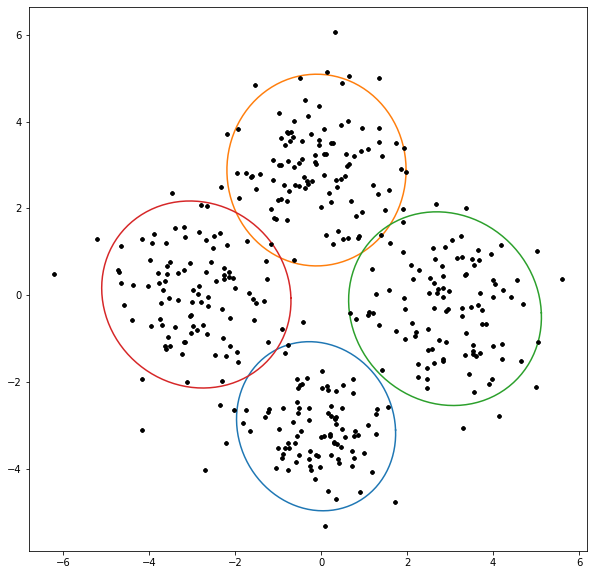} 
        \caption{D.o.F = 10}
    \end{subfigure}
    
	\caption{ SIA improves the clustering performance when the Gaussian components are corrupted by a multivariate t-distribution - Refer to table \ref{table:T_dist_sia_improv}}
	\label{fig:corruption_sia_improved}
\end{figure*}

\begin{table*}[!htb]
\centering
\captionof{table}{ Improvement in clustering performance before and after SIA step-2 for the contaminated case}
\label{table:T_dist_sia_improv}
\begin{tabular}{c|p{1.35cm} p{1.25cm}p{1.25cm}p{1.25cm}p{1.25cm}|p{1.35cm}p{1.25cm}p{1.25cm}p{1.25cm}p{1.25cm} } 
 \hline
  & \multicolumn{5}{c|}{Before SIA Step 2}  & \multicolumn{5}{c}{After SIA Step 2}   \\ \hline
 DoF & LL & KLF & KLB & MPKL & ARI  & LL & KLF & KLB & MPKL & ARI \\ \hline
 
 2 & -1837.6 & 158.4 & 60.7 & 58.6 & 0.57 & -1943.9 & 36.8 & 41.6 & 8.8 & 0.76 \\ %
 3 & - 1813.4 & 57.2 & 80.9& 16.6& 0.37 & -1871.0 & 48.2 & 46.2 &4.3 &0.79 \\ %
 4 & -1725.8 & 48.6 & 89.1 & 11.7 & 0.75 & -1735.1 & 39.6 & 54.3 & 3.8 & 0.78 \\ %
 5 & -1683.9 & 67.1 & 82.9 & 11.12 & 0.79 & -1694.5 & 46.1 & 54.6 & 4.9 & 0.82 \\ %
 10 & - 1640.7 & 68.0 & 96.9 & 13.9 & 0.87 & -1648.3 & 56.3 & 65.8 & 4.2 & 0.87 \\ \hline 
\end{tabular}
\end{table*}

On the high-dimensional dataset, from section \ref{sec:autodiff} we find that SIA increases the variance of the non-dominating component (fig. \ref{fig:sia2_hd} (a)) and the clustering performance remains low with ARI = 0.048.
Running SIA-HD gives an ARI of 0.35; furthermore, there is no dominating component in the final clustering (fig. 
\ref{fig:sia2_hd} (b)).

Fig. \ref{fig:corruption_sia_improved} shows the clustering obtained by SIA on the contaminated mixture data from section \ref{sec:penalty}.
Table \ref{table:T_dist_sia_improv} shows the values of Loglikelihood (LL), ARI, MPKL, KLF and KLB before and after step 2 of SIA. 
The use of SIA decreases the likelihood but improves the clustering performance (ARI).
The improvement in ARI is higher for the case with smaller degree of freedom, where misspecification is higher. A similar analysis shows the efficacy of SIA on datasets contaminated with random noise in Appendix \ref{section:appendix_noise}.

A limitation of SIA is that when cluster covariances are asymmetric in terms of their volume, its performance  deteriorates. For a discussion on clustering such data we refer the reader to \cite{aitkin1980mixture,jorgensen1990influence}.

\section{Simulation Studies}
\label{sec:sim}

\subsection{Misspecified Mixtures}

We simulate non-Gaussian datasets with varying cluster separation and imbalance across component sizes.
We evaluate the clustering performance of misspecified 3-component GMM  
using three different inference techniques -- EM, AD-GD and our SIA algorithm.

\subsubsection*{Varying Cluster Separation}
We simulate 50 datasets from a 
2-dimensional mixture model with 3 Gaussian components,
$\mathcal{N}((\lambda,\lambda),\mathbb{I}_2)$, 
$\mathcal{N}((-\lambda,\lambda),\mathbb{I}_2)$, 
$\mathcal{N}((\lambda,-\lambda),\mathbb{I}_2)$,
where  $\lambda$ is a scaling factor which controls the cluster separation 
and $\mathbb{I}_2$ is a unit covariance matrix.
We evaluate the performance for $\lambda$ values $\{3,4,5,7\}$.
We sample 100 datapoints from each of these 3 components. 
These sampled datapoints are cubed
so none of the components is normally distributed. 
  
Table \ref{table:simul_LD} shows the mean ARI, with standard deviation (SD), over these datasets, of the clusterin solutions obtained by the algorithms.
We observe that only when there is high cluster separation ($\lambda = 7$), EM performs well. 
In the remaining three cases, of lower cluster separation,  
SIA outperforms AD-GD and EM.

	\begin{table}[!h]
	\centering
\captionof{table}{ARI (mean and SD) on varying $\lambda$}
\label{table:simul_LD}
\begin{tabular}{ c p{1cm}p{1cm}p{1cm}p{1cm} } 

 \hline
 $\lambda$ &  \text{3} & \text{4}  & \text{5} & \text{7}\\ \hline
SIA & \textbf{0.517 \newline (0.091)} & \textbf{0.812 \newline(0.093)}& \textbf{0.969 \newline(0.039)} & 0.994 \newline(0.007) \\ 
AD-GD & 0.311 \newline (0.211) & 0.686 \newline(0.122) & 0.856 \newline (0.076) & 0.997 \newline(0.004)\\
EM & 0.162 \newline (0.112) & 0.416 \newline(0.061) & 0.864 \newline (0.175) & \textbf{1 \newline(0)}\\

 \hline
\end{tabular}
\end{table}

\subsubsection*{Unbalanced mixtures}
We simulate data from a 2 dimensional 2-component mixture. The mixture means $(\bmu_1,\bmu_2)$ are {((0.5,0),(-0.5,0))} and covariances matrices are $\bSigma_k = (\bSigma_k^{1/2} (\bSigma_k^{1/2})^T )$. The parameters of $\bSigma_k^{1/2}$ are sampled randomly to capture different covariance structures. The simulated datapoints are then cubed (for misspecification). 
Keeping the number of data points ($N_1$) in cluster 1 constant at 100, we vary the number of datapoints ($N_2$) in cluster 2 as $\{ 20,50,100 \}$. We simulate 15 different datasets for each setting.

The mean ARI obtained by the methods are given in table \ref{table:simul_unbal}. 
We observe that SIA outperforms AD-GD and EM  in all cases.
Appendix \ref{app:unbalanced} shows an illustration on a specific dataset.

\begin{table}[!h]
\centering
\captionof{table}{ARI (mean and SD) on varying imbalance }
\label{table:simul_unbal}
\begin{tabular}{c p{1 cm}p{0.75cm}p{0.75cm} } 

 \hline
 $N_2$ &  \text{100} & \text{50}  & \text{20}\\ \hline
\text{SIA} & \textbf{0.123 \newline(0.157)} & \textbf{0.195 \newline(0.247)} & \textbf{0.224 \newline(0.286)} \\
\text{AD-GD} & 0.101 \newline (0.180) & 0.148 \newline (0.239) &0.151 \newline (0.275) \\  
\text{EM} & 0.101 \newline (0.130) & 0.150 \newline (0.221) &0.196 \newline (0.282) \\

 \hline
\end{tabular}

\end{table}

\subsection{High-Dimensional Data}

We compare the performance of EM and AD-GD (on both GMM and MFA) with that of SIA-HD, on high-dimensional datasets with varying $n/p$ and covariance structures.

\subsubsection*{Varying $\mathbf{n/p}$}
Following \citep{hoff_2004_jrssb,hoff_2005_biometrics, pan2007penalized,wang2008variable,guo2010pairwise}, we simulate 10 datasets with 100 datapoints 
from a 2-component GMM, with 50 datapoints from each component. In these datasets both the components have unit spherical covariance matrices
but differ in their mean vectors. 
90\% of the dimensions of the mean vector are kept same (with value 0) across both the components and 10\% of the dimensions of the mean vector differ (0 in one and 1 in the other cluster).
Hence, for the purpose of clustering, only 10\% of the dimensions are discriminating features and remaining 90\% of the dimensions are noise features. 

Table \ref{table:simul_HD} shows the mean ARI obtained by SIA-HD, AD-GD and EM by fitting GMMs and MFA on high-dimensional datasets.
EM fails in high dimensions while all the AD-GD algorithms run to completion.
In the more challenging cases of low $n/p$ ratios SIA-HD outperforms AD-GD for both GMM and MFA.

\begin{table}[!h]
\centering
\captionof{table}{ARI (mean and SD) on varying n/p }
\label{table:simul_HD}
\begin{tabular}{cp{0.95cm}p{0.95cm}p{0.95cm}p{0.95cm} } 

 \hline
 n/p &  \text{100/200} & \text{100/100}  & \text{100/50} & \text{100/10}\\ \hline
SIA-HD (GMM) & \textbf{0.801 \newline (0.058)} & \textbf{0.517 \newline (0.131)} & \textbf{0.333 \newline (0.168)} & 0.046 \newline(0.045) \\ 
AD-GD (GMM) & 0.738 \newline (0.051) & 0.474 \newline (0.135) & 0.239 \newline (0.146) & \textbf{0.046 \newline(0.040)} \\

SIA-HD (MFA) & 0.753 \newline (0.086) & 0.330 \newline (0.114) & 0.250 \newline (0.180) & 0.015 \newline(0.030)  \\
AD-GD (MFA)& 0.523 \newline (0.119) & 0.275 \newline (0.069) & 0.124 \newline (0.113) &  0.018 \newline(0.031)\\
EM (GMM) & - & - & - & 0.028 \newline(0.041) \\
EM (MFA) & - & - & 0.098 \newline (0.128) &  0.006 \newline(0.023)  \\

 \hline
\end{tabular}
\end{table}

\subsubsection*{Varying Covariance}
To evaluate the effects of  various covariance structures, we sampled two $p$-dimensional mean vectors, $\bmu_1,\bmu_2$, and two $p \times p$ matrices, $\bSigma^{\frac{1}{2}}_1, \bSigma^{\frac{1}{2}}_2$. Each element of the vectors $\bmu_k$ and $\bSigma_k$ are sampled independently from a standard normal distribution.  In order to ensure the covariance matrix is PD, we multiply the sampled matrices with their respective transposes, i.e. $\bSigma_k = \bSigma^{\frac{1}{2}}_k \bSigma^{\frac{1}{2}^{T}}_k$. 
We simulate 50 datasets for each case.

Table \ref{table:app_hd_simulations} shows that SIA-HD outperforms AD-GD, with the improvement increasing with decreasing $n/p$.
Only the best performing baseline method (AD-GD) is shown. 

\begin{table}[!h]
\begin{center}
\captionof{table}{ARI (mean and SD) for varying covariance structures}
\label{table:app_hd_simulations}
\begin{tabular}{cp {1 cm} p{1cm} p{1cm} } 
 \hline
 {\color{black}n/p} & \text{100/200} & \text{100/100} & \text{100/50}   
 \\ \hline
SIA-HD (GMM) &  \textbf{0.1825 \newline(0.210)}  & \textbf{0.256 \newline(0.241)}  & \textbf{0.108 \newline (0.140)} \\ 
AD-GD (GMM)  & 0.079 \newline(0.119)  &  0.124 \newline(0.185) & 0.108 \newline (0.145) \\ 

 \hline
\end{tabular}
\end{center}
\end{table}

\subsubsection*{Model Selection in High Dimensions}
We follow the approach used in \citep{hoff_2004_jrssb,hoff_2005_biometrics,pan2007penalized,wang2008variable,guo2010pairwise} and simulate a 4 component 50-dimensional dataset as shown in table \ref{table:clusterselc_data}. 
We assume unit spherical covariance matrices for each component. 
Further, the parameters of each component across the 50 dimensions are varied as shown in table \ref{table:clusterselc_data}, i.e., all the components have the same parameters for 35 dimensions, for component 2 only dimensions 1-5 are discriminating, for component 3 only dimensions 5-10 are discriminating and for component 4 only dimensions 10-15 are discriminating. 
The value of $\lambda$ controls the cluster separation. 
For each component we sampled ten datapoints.  
Three different sets of simulated data with varying cluster separation are obtained by choosing the value of $\lambda$  to be 1, 5 and 10 respectively. 
This experiment is repeated with 10 different seeds for each value of $\lambda$. 
We use SIA-HD for clustering with each value of $K \in \{3,4,5\} $ .
We compare the number of clusters selected using AIC and MPKL. 

\begin{table}[h]
\centering
\captionof{table}{Simulations for Model Selection}
\label{table:clusterselc_data}
\begin{tabular}{c p{1.1cm}p{1.1cm}p{1.1cm}p{1.1cm} } 

 \hline
 Features &  \text{C-1} & \text{C-2}  & \text{C-3} & \text{C-4} \\ \hline
1-5 & $\mathcal{N}(0,1)$ & $\mathcal{N}(\lambda,1)$  & $\mathcal{N}(0,1)$ & $\mathcal{N}(0,1)$ \\ 
5-10 & $\mathcal{N}(0,1)$ & $\mathcal{N}(0,1)$ &  $\mathcal{N}(\lambda,1)$  & $\mathcal{N}(0,1)$ \\ 
10-15 & $\mathcal{N}(0,1)$  & $\mathcal{N}(0,1)$ & $\mathcal{N}(0,1)$ & $\mathcal{N}(\lambda,1)$ \\

15-50 & $\mathcal{N}(0,1)$ & $\mathcal{N}(0,1)$ & $\mathcal{N}(0,1)$ & $\mathcal{N}(0,1)$ \\

 \hline
 
\end{tabular}
\end{table}

\begin{table}[!h]
\centering
\captionof{table}{ Number of times 3, 4, and 5 components are selected using each criterion for a 4-component dataset.}
\label{table:clusterselc_results}
\begin{tabular}{c |c |p{1cm}p{1cm}p{1cm} } 

 \hline
 $\lambda$& \# clusters: &  \text{3} & \text{4}  & \text{5}  \\ \hline
1 & MPKL &  10  & 0 & 0 \\ 
1 & AIC &  10 &  0  & 0 \\
 1 & BIC &  10 &  0  & 0 \\ \hline
5 & MPKL &  1 & \textbf{7} & 2 \\
5 & AIC &  10 &  0  & 0 \\
 5 & BIC &  10 &  0  & 0 \\ \hline
10 & MPKL &  1 & \textbf{8} & 1 \\ 
5 & AIC &  0 &  10  & 0 \\
 10 & BIC & 10 &  0  & 0 \\
 \hline
 
\end{tabular}
\end{table}
Table 
\ref{table:clusterselc_results}
shows that at very low cluster separation ($\lambda=1$), both the criteria do not select 4 clusters.
At moderate and high cluster separation ($\lambda=5, 10$), BIC underestimates the number of clusters to 3, which is consistent with previous findings %
\citep{melnykov2010finite}. AIC, on the other hand, performs well only when the cluster separation is very high.
MPKL identifies 4 clusters in 7 out of 10 times at moderate separation and 8 out of 10 times at high cluster separation.

\section{Real Datasets}
\label{sec:real}

We compare the performance of SIA to AD-GD and EM on five real datasets -- 
Wine \citep{Forina:etal:1986}, 
IRIS \citep{fisher1936use}, 
Abalone \citep{nash1994population}), 
Urban Land cover \citep{johnson2013classifying} and LUSC RNA-Seq \citep{kandoth2013mutational}). 
\begin{table}[!h]
	\caption{Real datasets: Clustering performance (ARI) of SIA, EM and AD-GD}
	\label{table:realdatasets}
	\begin{center}
		\begin{tabular}{ c|ccc|ccc } 
			\hline
			Dataset & n & p & K & SIA & EM  & AD-GD  \\ \hline
			\text{IRIS}&150 & 4 &3 &  0.922 & 0.903 & 0.903  \\ %
			\text{Wine}&178 & 13 & 3 & 0.570 & 0.462 & 0.375   \\ %
			\text{Abalone}&4177 & 8& 3& 0.128 & 0.121 & 0.089  \\ %
			\text{Urban} & 168& 148& 9 &0.112 & - & 0 \\ %
			\text{LUSCRNAseq } & 130& 206 & 2 & 0.0228 & - & 0  \\ \hline
		\end{tabular}
	\end{center}
	
\end{table}

Table \ref{table:realdatasets} shows the number of observations and dimensions in each dataset.
Further, using the number of clusters and ground truth labels provided, we compare the ARI.
We observe that SIA obtains the highest ARI in all the five datasets.
In high-dimensions, EM fails to run and AD-GD  assigns all the points to a single cluster corresponding to the dominating component.
We now present detailed analysis of Wine and LUSCRNA-Seq datasets.

\subsection{Wine dataset}

\begin{table*}[!h]
\centering
\captionof{table}{ Clustering Outputs for Wine dataset}
\label{tab:winedataset}
\begin{tabular}{c c p{1.5cm}p{1.5cm}p{1.5cm}p{1.5cm}p{1.5cm}p{1.5cm}p{1.5cm}  } 
 \hline
S. No.  & $K$ &  Algorithm &  LL & AIC & KLF & KLB & MPKL & ARI  \\ \hline
1 & 2 & EM & -2983.6 & 6385.3 & 86.3 & 39.5 & 46.8 & 0.46 \\ %
 2 & 2 & EM & -3117.4 & 6652.9 & 17.2 & 14.8 & 2.3 & 0.57 \\ %
 3 & 2 & EM & -3083.0 & 6584.0 & 22.2 & 17.36 & 4.8 & 0.59 \\ %
 4 & 2 & AD-GD & -3081.6 & 6581.2 & 56.1 & 29.6 & 26.5 & 0.55 \\ %
 5 & 2 & SIA & -3072.0 & 6562.0 & 42.1 & 27.5 & 14.6 & 0.57 \\ %
 \hline
 6 & 3 & EM & -2901.0 & 6430.0 & 321.29 & 91.44 & 229.8 & 0.46 \\ %
 7 & 3 & EM & -2915.7 & 6459.4 & 93.25 & 118 & 51.68 & 0.618 \\ %
 8 & 3 & AD-GD & -3030.04 & 6688.8 & 165.9 & 152.1 & 50.7 & 0.385 \\ %
 9 & 3 & SIA & -2921.45 & 6470.9 & 57.7 & 60.8 & \textbf{14.66} & \textbf{0.630} \\ %
 \hline
 10 & 4 & EM & -2637.7 & \textbf{6113.4} & 28622214.3 & 407.7 & 13486984 & 0.52 \\ %
11 & 4 & EM & -2660.7 & 6158.0 & 5136379.7 & 362.1 & 5136017.5 & 0.52 \\ %
  12 & 4 & AD-GD & -2910.7 & 6659.4 & 1192.4 & 364.0 & 486.65 & 0.3546 \\ %
 13 & 4 & SIA & -2850.5 & 6539.0  & 519.0 & 304.9 & 214.05 & \textbf{0.67} \\ %
 \hline
\end{tabular}
\end{table*}

\begin{figure}[!h]
\centering
\includegraphics[width=0.3\textwidth]{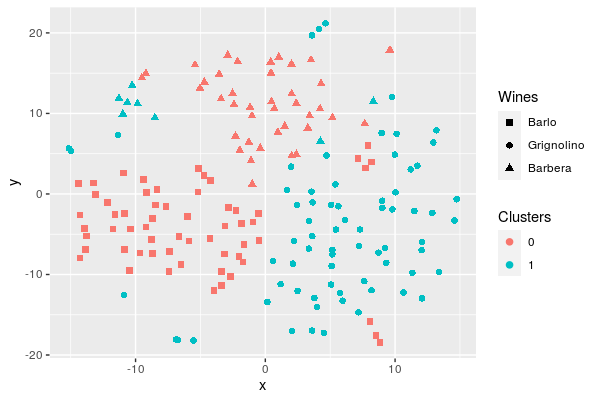}
\includegraphics[width=0.3\textwidth]{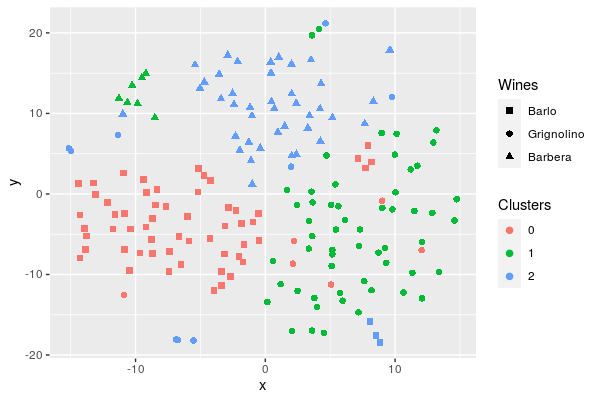}
\includegraphics[width=0.3\textwidth]{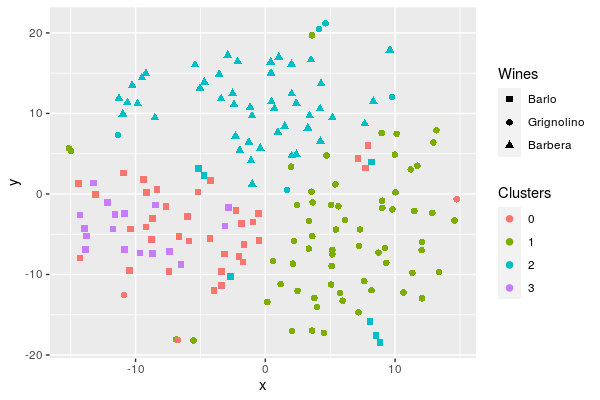}
\caption{tSNE plots for the best model for each value of $K$ 
(S. Nos. 3,9 and 13 in Table \ref{tab:winedataset});
Clusters obtained for $K=2$ (above), $K=3$ (middle) and $K=4$ (below).}
\label{fig:wine_tsne}
\end{figure}

\begin{table*}[!h]
\caption{Wine data: Clustering results for the best model for each $K$ (S. Nos. 3,9 and 13 in Table \ref{tab:winedataset})}
\label{tab:wine clustering}
\centering
\begin{tabular}{l c rr c rrr c rrrr }
\toprule
    && \multicolumn{2}{c}{$K = 2$} && \multicolumn{3}{c}{$K=3$} && \multicolumn{4}{c}{$K=4$} \\
 \cline{3-4}\cline{6-8}\cline{10-13}
Cultivar && 1 & 2  && 1 & 2 & 3  && 1 & 2 & 3 & 4 \\  
\midrule
 Barolo    && 59 &       && 56 &    & 3         && 35 &    &7    &17       \\ 
 Grignolino    &&    & 71     &&  6  & 56 & 9      &&  3  & 63 & 5   &     \\ 
Barbera   &&  40 &  8   &&   & 7   & 41   &&   &    & 48 &   \\ 
\bottomrule
\end{tabular}
\end{table*}

\begin{figure*}[!h]
    \centering
    \includegraphics[width=0.7\textwidth]{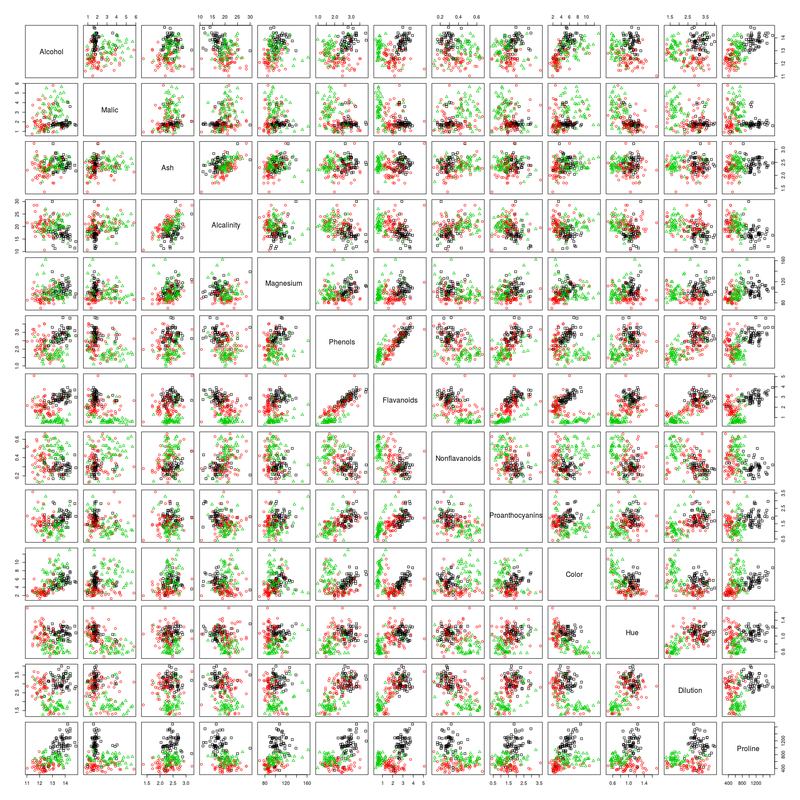}
    \caption{Scatterplot matrix for the recommended model}
    \label{fig:wine_scatter}
\end{figure*}

The Wine dataset \citep{forina1991uci} contains the results of a chemical analysis of wines grown in the same region in Italy but derived from three different cultivars. The analysis determined the quantities of 13 constituents found in each of the three types of wines. We use this dataset, without the labels of wine types, for analyzing the performance of the clustering algorithms.

We fit EM, AD-GD and SIA for $K \in \{ 2,3,4 \}$ and report their log-likelihood (LL), MPKL, KLDivs, AIC and ARI values in table \ref{tab:winedataset}. For a given value of $K$ we show the results for multiple runs of EM with different initializations
to illustrate how loglikelihood along with our proposed MPKL heuristic can be used for model selection.

For $K=2$, among the three solutions from EM - while clustering output (S. No. 1) has better LL of 2983.6, its MPKL of 46.8 is higher (worse) than those obtained by the other clustering outputs (S. No. 2 and S. No. 3). So, even among EM solutions, by trading off likelihood for an improvement in MPKL, we can choose better clustering solutions with higher accuracy. 
The LL of SIA is better than that of AD-GD (which was used in step 1 of SIA) indicating the KLDivs based penalization can aid in escaping local maxima. 
The ARI obtained by SIA (S. No. 5) is comparable to the best results from EM.

We observe similar trends for $K=3$.
Among EM solutions (S. No. 6 and 7), the solution with better MPKL (S. No. 7) achieves a better ARI.
Again, the LL of SIA (S. No. 9) is better than that of AD-GD (S. No. 8 which was used in step 1 of SIA) indicating the KLDivs based penalization can escape local maxima. The best ARI is achieved by SIA which also has the lowest MPKL value.

For $K=4$, we observe that AD-GD solution (S. No. 12) does not have high LL, nor does it have a low MPKL value. So, we initialize SIA step-2 with the output of EM; we observe a decrease in MPKL and an improvement in the clustering performance (ARI) over both EM and AD-GD. 

Given the low MPKL values and better AIC values, we recommend S. No. 9 as the best model for this dataset. 
We give the confusion matrices and tSNE plots \citep{maaten2008visualizing} of the best (with respect to the MPKL heuristic) models for each $K$ in Table \ref{tab:wine clustering} and in fig. \ref{fig:wine_tsne} respectively. From the tSNE plots as well, we can see that $K=3$ gives best and most well separated clusters. The scatterplot matrix for the recommended model is given in fig. \ref{fig:wine_scatter}.

\subsection{Lung Cancer RNA-Seq Dataset}
The LUSCRNA-seq data \citep{kandoth2013mutational}
contains 
gene expression and methylation data
(206 attributes in total) for 130 patients with lung cancer
as available from the 
TCGA database \citep{weinstein2013cancer}. 
Following \citet{kasa2019gaussian}, 
we use the (binary) survival status, which was not used as input for clustering, as the labels for computing ARI.

The MPKL criterion identifies 2 clusters as shown in Table \ref{tab:mpkl_luscrna}.
\begin{table}[!h]
\begin{center}
\captionof{table}{MPKL for LUSCRNA-seq dataset }
 \label{tab:mpkl_luscrna}
\begin{tabular}{cccc } 
 \hline
 \text{\# of clusters} & 2 & 3 & 4 
 \\ \hline
 \text{MPKL} & 6.739 & 27.26 & 9.74  \\ \hline
\end{tabular}
\end{center}
\end{table}
We compare the performance of SIA-HD with EM (MFA) and 
state-of-the-art model-based clustering algorithms for high-dimensional data:
\begin{itemize}
\item
HDDC \citep{bouveyron2014model}, based on subspace clustering and parsimonious modeling.
\item
VarSelLCM \citep{marbac2017variable}, based on variable selection.
\item
 MixGlasso \citep{stadler2017molecular}, a penalized likelihood approach designed to automatically adapt to the sample size and scale of clusters.
\end{itemize} 

\begin{table}[!h]
 \begin{center}
\captionof{table}{Clustering performance on LUSCRNA data}
\label{table:ari_luscrna}
\begin{tabular}{ccc } 
 \hline
Algorithm & \text{ARI} & \text{RI}%
 \\ \hline
SIA-HD (MFA) & 0.0228 & 0.51\\ %
EM (MFA) & - & -\\ 
HDDC & - & -  \\ 
MixGlasso & -0.01 & 0.49 \\
VarSelLCM & -0.08 & 0.49 \\
 \hline
\end{tabular}
\end{center}
\end{table}

\begin{figure}[!h]
\centering
\includegraphics[width=0.35\textwidth]{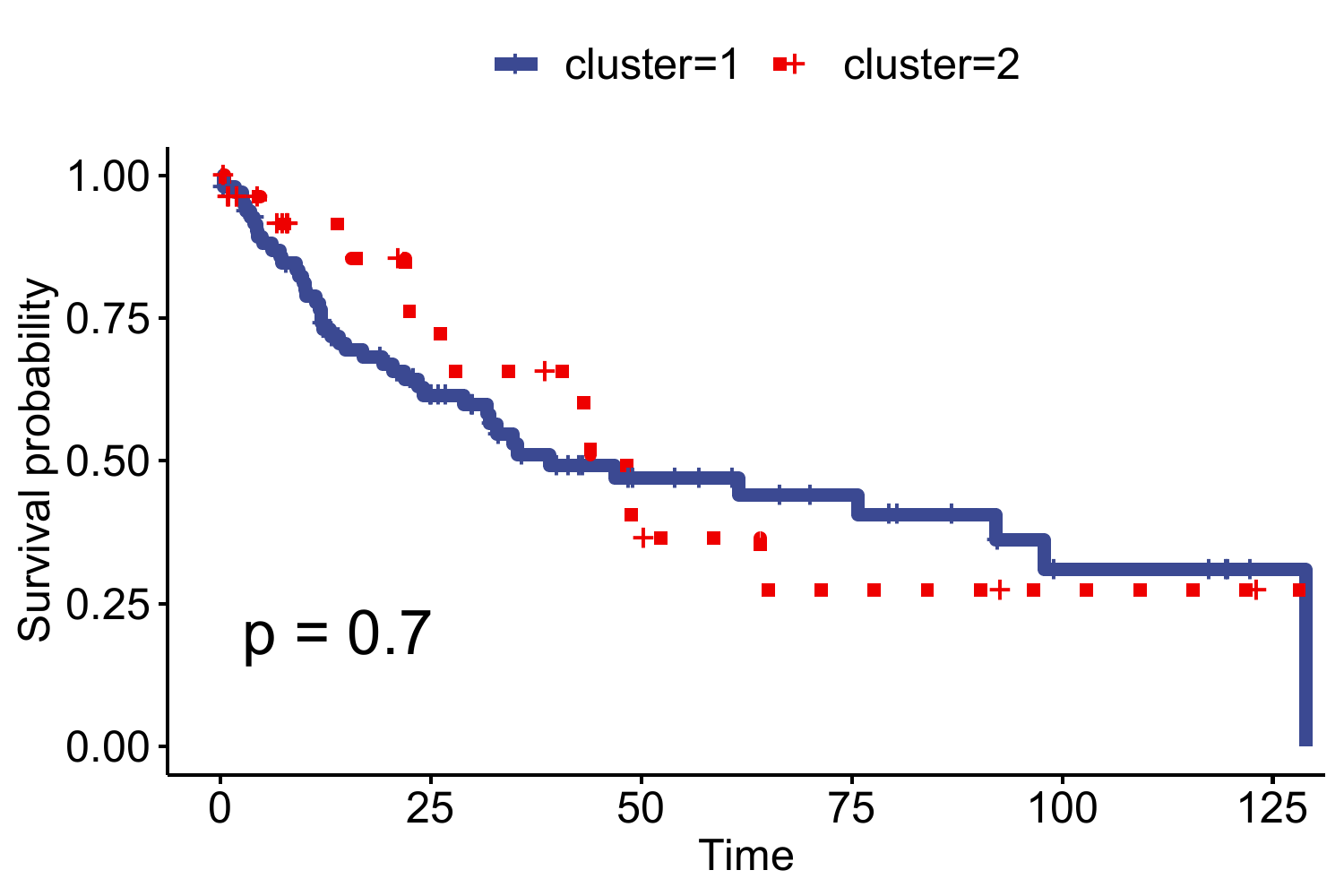}
\includegraphics[width=0.35\textwidth]{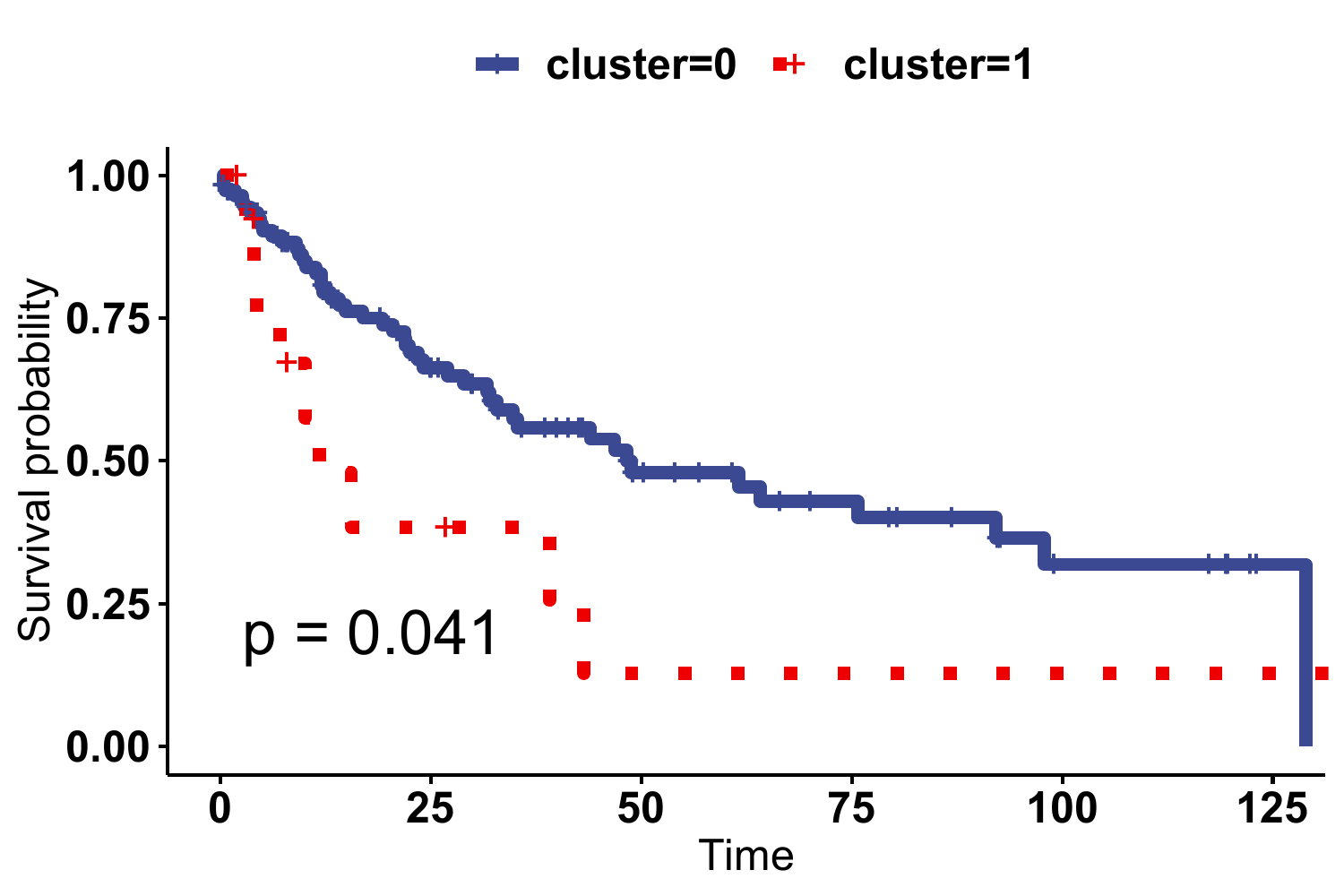}
\caption{Kaplan-Meier survival curves with $p$-values; %
Clusters obtained from MixGlasso (left) and SIA-HD (right).}
\label{fig:survival}
\end{figure}
\begin{figure}[!h]
     \centering
     
    \begin{subfigure}[b]{0.365\textwidth}
\includegraphics[width= \textwidth]{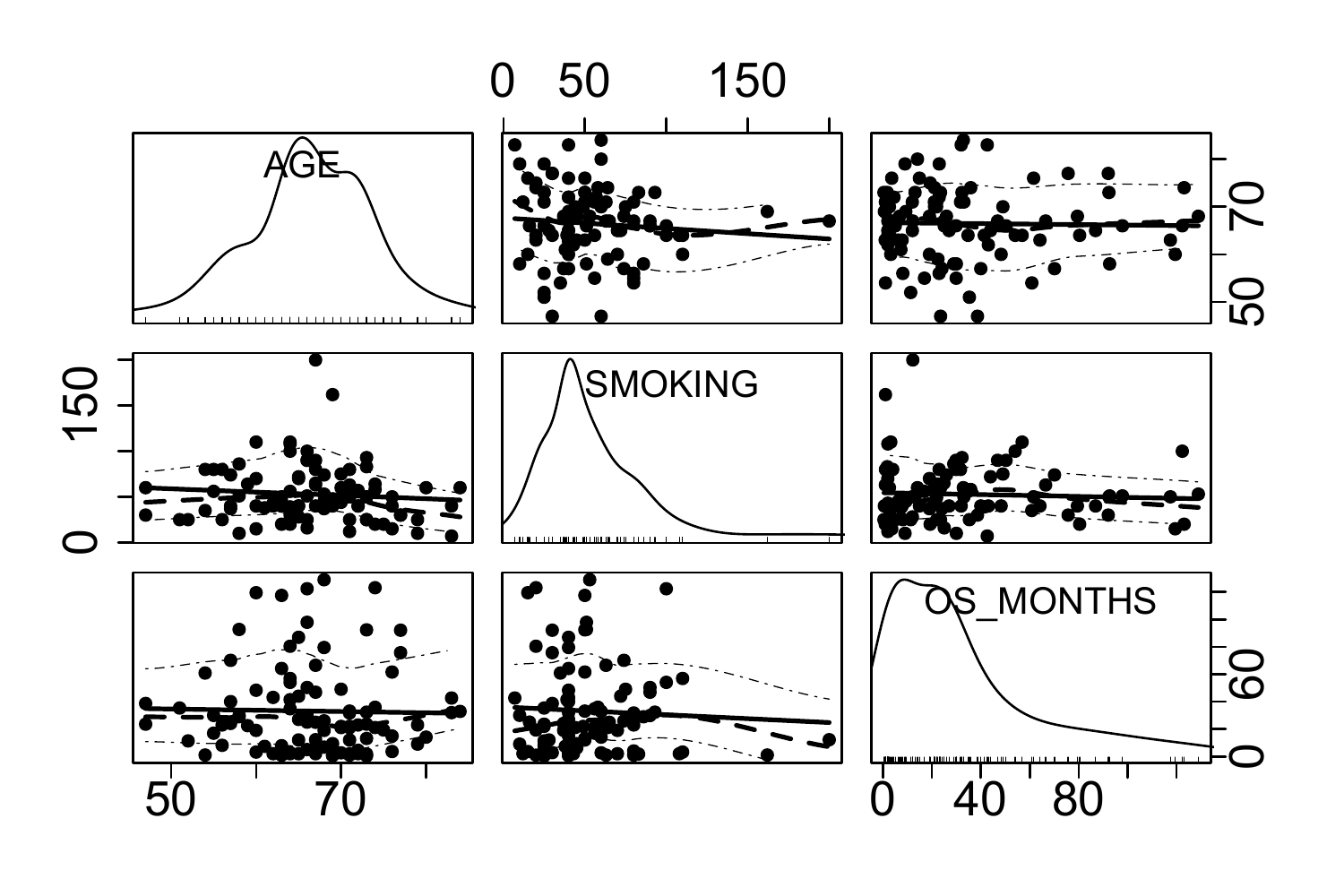}
        \caption{High Survival }
    \end{subfigure}
    \begin{subfigure}[b]{0.365\textwidth}
\includegraphics[width= \textwidth]{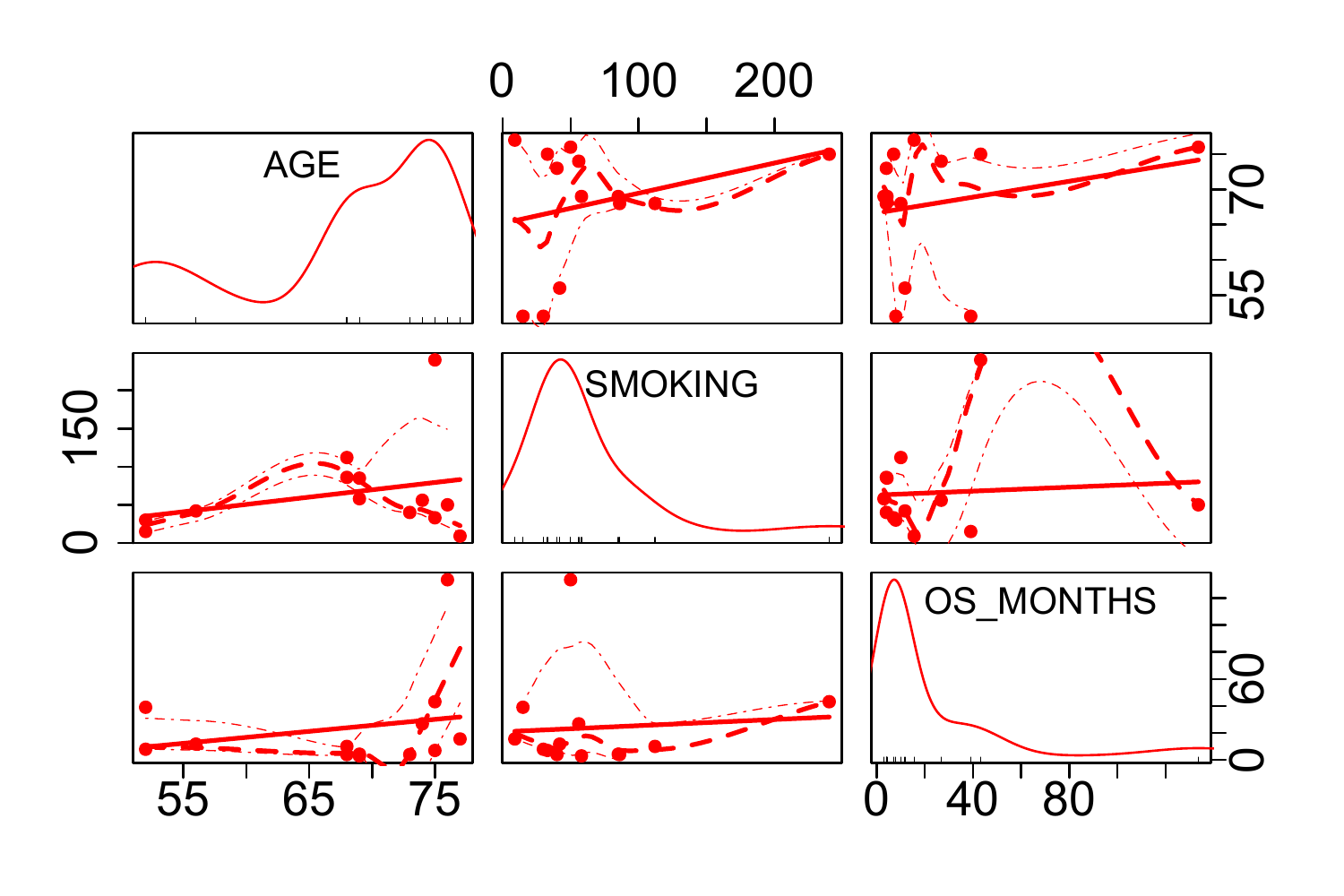}
        \caption{ Low Survival }
    \end{subfigure}%

 \caption{Bivariate scatterplots of three features -- age (in years), smoking (number of packs per year) and survival (in months)  -- for two clusters obtained from SIA-HD.}
 \label{fig:scatter}
 \end{figure} 
 
For all EM-based algorithms, 5 different K-Means initializations were used, where
K-Means itself uses a random initialization.
The result with the best AIC is reported.
Table \ref{table:ari_luscrna} shows the ARI and Rand Index (RI)
obtained by all the methods. 
EM and HDDC fail to run. 
SIA-HD has the best ARI and RI.
 
We  evaluate the clusters obtained 
by SIA-HD and MixGlasso,
with respect to their clinical subtypes, in terms of survival probability. 
The survival probabilities are estimated using Kaplan-Meier method and  
the $p$-values are obtained by the log-rank test on the survival analysis based on the cluster memberships.
Fig. \ref{fig:survival} shows the survival curves obtained on clusters from SIA-HD and MixGlasso.
We observe that only SIA-HD yields significant clusters ($\textit{p-} \text{value} < 0.05$) that
are the least overlapping indicating a good separation of patient subtypes.  
 
Fig. \ref{fig:scatter} shows bivariate dependencies in the two clusters from SIA-HD, thereby illustrating its ability to find meaningful patterns. Each scatterplot shows pairwise dependencies and the univariate distribution (diagonal). The bivariate pattern between smoking and survival is distinctly different between the two clusters: in the cluster with higher survival probability patients tend to smoke less when they are older.

\section{Conclusion}
\label{sec:concl}

In this paper we empirically investigated model-based clustering using EM and AD-GD in two practically relevant cases of misspecified GMMs and high-dimensional data.
Empirically we found that EM has better clustering performance (in terms of ARI) in cases of misspecification 
while AD-GD outperforms EM on high-dimensional data. 
However, in misspecified cases, we observe inferior clustering solutions with poor ARI both from EM and AD.
Such solutions have components with large variance and occur frequently with many different initializations.
With high-dimensional data, we find that inferior solutions have components %
with low variance and high mixture weight.

To address these problems, we proposed a new penalty term based on the Kullback Leibler divergence between pairs of fitted components.
We
developed algorithms SIA, SIA-HD for clustering and the 
MPKL criterion for model selection.
Gradient computations for this penalized likelihood is difficult but AD-GD based optimization, that does not require closed-form derivatives, can be done effortlessly.
Experiments on synthetic and real datasets demonstrate the efficacy of this penalized likelihood for clustering and model selection -- in cases of misspecification and high-dimensional data.
Software implementation of our algorithm and the experiments conducted are available
online\footnote{\url{https://bitbucket.org/cdal/sia/}}.

Our experiments also illustrate the benefits of automatic differentiation (AD), that is well known but rarely used, for parameter inference in mixture models.
We compared AD-GD with EM for two cases -- GMM with unconstrained covariances and MFA.
The flexible framework of AD-GD and our KL divergence based penalty term 
can be used for many families of mixture models, e.g., PGMM, as constraints imposed by these models can be easily added to AD-GD.
A limitation of SIA is that when cluster covariances are asymmetric in terms of their volume, its performance  deteriorates.
Future work can address this limitation and explore the 
use of our general KL divergence based penalty term for non-Gaussian mixtures and our model selection criterion for other models.
Theoretical properties such as convergence and sample complexity can also be investigated in the future.

\begin{acknowledgements}
V.R. was supported by Singapore Ministry of Education Academic Research Fund [R-253-000-138-133].
\end{acknowledgements}

\bibliographystyle{spbasic}      %
\bibliography{main}

\appendix
\onecolumn
\section{Symbols and Notation}
\label{sec:symbols}

	\begin{table}[!h]
	\centering
\begin{tabular}{cc }%

 \hline
 Symbol &  Meaning  \\ \hline
 $f$ & overall probability density from which data is sampled\\
 $g$ & misspecified model for data sampled from $f$ \\
 $f_i$ & probability density of the $i$-th component \\
 $\pi_i$ & mixing proportions or weight of the $i$-th component \\
 $\btheta_i$ &  parameters of $i$-th mixture component \\
 $\btheta$ & all parameters of the model\\
 $\btheta^{*}$ & true parameters of the model from which data is sampled \\
 $\bTheta$ & set of all possible parameters i.e. the entire parameter space \\
 $\bF_i$ & factors associated with $i$-th datapoint \\
 $K$ & total of number of components in the model \\
 $k,k^{'},k_1,k_2$ & indices over the total number of components $K$  \\
 $\bX_i$ & $i$-th random datapoint \\
 $\bx_i$ & $i$-th observed datapoint \\
 $\bX$ & all the $n$ random datapoints \\
 $\bx$ & all the $n$ observed datapoints \\
 $n$ & Total number of $iid$ datapoints considered \\
 $p$ & Total number of dimensions or features of the data \\
 $p_e$ & Total number of free parameters in the model \\
 $\mathcal{L(\btheta)}$ & Log-Likelihood of observing $\bx$ if the parameters were $\btheta$\\ 
 $\mathcal{N}$ & density of multi-variate Gaussian distribution \\
 $\bmu_i, \mu_{ij} $ & mean of $i$-th component of a GMM and its value of along $j$-th dimension respectively \\
 $\bSigma_i, \Sigma_{ijk}$ & Covariance matrix of $i$-th component of a GMM and its $jk$-th element respectively \\
  $\bU_i$ & square-root of $i$-th covariance matrix $\bSigma_i$\\
  $\hat{a}^{t}$ &  Estimate of parameter $a$ at the end of iteration $t$ \\ 
  $c,C$ & Constants independent of model parameters \\ 
  $\lambda$ & factor controlling the cluster separation in simulations \\ 
  $\lambda_k$ & hyperparameter controlling the fitted component covariance \\ 
  $\mathbb{I}$ & Identity matrix \\ 
$\epsilon$ & learning rate in the vanilla gradient descent step \\
$\gamma$ & convergence threshold for stopping criterion \\

 \hline
 
\end{tabular}
\captionof{table}{Symbols used in the paper }
\label{table:symbols}
\end{table}

\section{Automatic Differentiation}
\label{section:AD}

The examples in this section are from \cite{baydin2018automatic}.
In symbolic differentiation, we first evaluate the complete expression and then differentiate it using rules of differential calculus. 
First note that a naive computation may repeatedly evaluate the same expression multiple times, e.g., consider the rules:
\begin{align}
\frac{d}{dx} \left(F(x) + G(x)\right) \leadsto \frac{d}{dx} F(x) + \frac{d}{dx} G(x) \\
\frac{d}{dx} \left(F(x)\,g(x)\right) \leadsto \left(\frac{d}{dx} F(x)\right) G(x) + F(x) \left(\frac{d}{dx} G(x)\right)
    \label{EquationMultiplicationRule}
\end{align}

Let $H(x)=F(x)G(x)$. 
Note that $H(x)$ and $\frac{d}{dx}H(x)$ have in  common: $F(x)$ and $G(x)$, 
and on the right hand side, $F(x)$ and $\frac{d}{dx}F(x)$ appear separately. 
In symbolic differentiation we plug the derivative of $F(x)$ and thus 
have nested duplications of any computation that appears in common between $F(x)$ and $\frac{d}{dx}F(x)$. 
In this manner symbolic differentiation can produce exponentially large symbolic expressions which take correspondingly long to evaluate. This problem is called \textbf{expression swell}.
To illustrate the problem, consider the following  iterations of the logistic map $l_{n+1}=4l_n (1-l_n)$, $l_1=x$ and the corresponding derivatives of $l_n$ with respect to $x$. Table \ref{TableExpressionSwell} clearly shows that the number of repetitive evaluations increase with $n$.

\begin{table}
  \centering
  \captionof{table}{Iterations of the logistic map $l_{n+1}=4l_n (1-l_n)$, $l_1=x$ and the corresponding derivatives of $l_n$ with respect to $x$, illustrating expression swell 
  (from \cite{baydin2018automatic}) }
  \label{TableExpressionSwell}
  \renewcommand{\arraystretch}{1}
  
  {\small
  \begin{tabularx}{\columnwidth}{@{}lXp{3cm}XX@{}}
    \toprule
    $n$ & $l_n$ & $\frac{d}{dx}l_n$ & $\frac{d}{dx}l_n$ (Simplified form)\\
    \addlinespace
    \midrule
    1 & $x$ & $1$ & $1$\\
    2 & $4x(1 - x)$ & $4(1 - x) -4x$ & $4 - 8x$\\
    3 & $16x(1 - x)(1 - 2 x)^2$ & $16(1 - x)(1 - 2 x)^2 - 16x(1 - 2 x)^2 - 64x(1 - x)(1 - 2 x)$ & $16 (1 - 10 x + 24 x^2 - 16 x^3)$\\
    4 & $64x(1 - x)(1 - 2 x)^2$ $(1 - 8 x + 8 x^2)^2$ & $128x(1 - x)(-8 + 16 x)(1 - 2 x)^2 (1 - 8 x + 8 x^2) + 64 (1 - x)(1 - 2 x)^2  (1 - 8 x + 8 x^2)^2 - 64x(1 - 2 x)^2 (1 - 8 x + 8 x^2)^2 - 256x(1 - x)(1 - 2 x)(1 - 8 x + 8 x^2)^2$ & $64 (1 - 42 x + 504 x^2 - 2640 x^3 + 7040 x^4 - 9984 x^5 + 7168 x^6 - 2048 x^7)$\\
    \bottomrule
  \end{tabularx}}
\end{table}

If the symbolic form is not required and only numerical evaluation of derivatives is required, computations can be simplified by storing the values of intermediate sub-expressions.
Further efficiency gains in computation can be achieved by interleaving differentiation and simplification steps.
The derivative of $l_{n+1} = 4 l_{n} (1- l_{n})$ can be found using the chain rule $\frac{dl_{n+1}}{dl_n}\frac{dl_{n}}{dl_{n-1}}\dots \frac{dl_{1}}{dl_x}$ which simplifies to $4(1-l_{n} - l_{n})4(1-l_{n-1} - l_{n-1})\dots4(1-x - x)$. 
Note that evaluation in AD is computationally linear in $n$ (because we add only one $(1-l_n - l_n)$ for each increase by 1).
This linear time complexity is achieved due to `carry-over' of the derivatives at each step, rather than evaluating the derivative at the end and substituting the value of x. 

\newpage
The Python code below shows the simplicity of the implementation for this problem.

\begin{center}
\begin{verbatim}
from autograd import grad
def my_func(x,n):
    p = x
    y = x * (1 - x)
    for i in range(n):
        y = y*(1 - y)
        
    return y
grad_func = grad(my_func)
grad_func(0.5,4)
\end{verbatim}
\end{center}

Consider the following recursive expressions: 
$l_0 = \frac{1}{1+e^x}$, $l_1 = \frac{1}{1+e^{l_0}} $, ....., $l_{n} = \frac{1}{1+e^{l_{n-1}}} $
We evaluate the derivative of $l_{n}$ with respect to $x$ and compare the runtime in Mathematica (SD) vs PyTorch (AD) for various values of $n$. As $n$ increases, it is expected that runtime also increases. It can be seen from the results in Table \ref{Runtime}  that runtime increases linearly for AD (using PyTorch) whereas it increases exponentially for SD (using Mathematica).

\begin{center}
\captionof{table}{Average runtime (over 1000 runs) }
\label{Runtime}
\begin{tabular}{ ccc } 

 \hline
 n & \text{AD}  & \text{SD} \\ \hline
1 & 0.00013 & 0.00000 \\ 
5 & 0.00030 & 0.00005 \\ 
10 & 0.00051 & 0.00023 \\
50 & 0.00293 & 0.00437 \\
100 & 0.00433 & 0.15625 \\
200 & 0.00917 & 1.45364 \\
 \hline
\end{tabular}
\end{center}

\section{Additional Experimental Results}

\subsection{EM and AD-GD: A comparison of clustering performance}
\label{app:robust_initialization}

We compare the 
clustering performance of 
AD-GD %
and EM on pure GMM data as well as non-GMM (misspecified) data. 
We do so by simulating 3600 datasets as described below and running both the algorithms with matching settings with respect to random initialization, number of iterations (maximum of 100 iterations)  and convergence threshold (1e-5).
The details of the simulations and results are given below. 

\paragraph{Pure GMM data}: We simulate datasets from 36 %
different 2--dimensional 3--component Gaussian Mixture Models that differ in their means and covariance matrices. 
This is to test different types of data -  from well-separated components to highly overlapping components. 
The GMM parameters are varied as follows:
\begin{itemize}
    \item First, we construct a $ 2 \times 1 $ vector $\bv_k$ and a $2 \times 2$ matrix $\bZ_k$ for each of these 3 components, where each element of the vectors and matrices are sampled from a standard normal distribution.
    \item The mean vector of the $k^{th}$ component is obtained by multiplying  $\bv_k$ by a factor $k_{\bmu}$ i.e. $\bmu_k = k_{\bmu} \bv_k$. This factor $k_{\bmu}$ is varied from 0.25 to 1.5 in steps of 0.25, i.e., the factor $k_{\bmu}$ takes one of the six values in \{0.25, 0.5, 0.75, 1, 1.25, 1.5\}. Higher the value of $k_{\bmu}$, more the separation between the means of components.
    \item The $\bZ_k$ vector is added to the $2 \times 2$ Identity matrix $\mathbb{I}$  (This step ensures the PD of the covariance matrix that is computed in the later steps).  
    \item We multiply the matrix $\bZ_k + \mathbb{I}$ by $k_{\bSigma}$ to obtain $\bU_k$. The scaling factor $k_{\bSigma}$ is varied from 0.05 to 0.65 in steps of 0.1, i.e., the scaling factor takes one of the six values in \{0.05, 0.15, 0.25, 0.35, 0.45, 0.55\}. The covariance matrix of the $k^{\rm th}$ component is $\bSigma_k = \bU_k \bU_{k}^T$. Higher the scaling factor $k_{\bSigma}$, more the inter-component overlap.
\end{itemize}

By choosing six different values of $k_{\bmu}$ and $k_{\bSigma}$ each, we simulate 36 (6x6) GMM parameters with varying cluster separation. For each these 36 GMM parameters, we simulate 50 different datasets each containing 300 samples, for a total of 1800 datasets. 
Both the methods (EM and AD-GD) are run on each dataset with 50 random initializations and average ARI is computed for each combination of $(k_{\bmu},k_{\bSigma})$.

\begin{figure}[H]
\centering

    \begin{subfigure}[b]{0.3\textwidth}
\includegraphics[width= \textwidth]{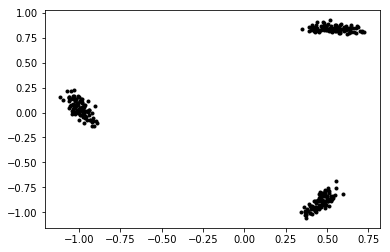}
        \caption{Radial - 0.05; Tangential - 0.05}
        
    \end{subfigure}%
    ~
 \begin{subfigure}[b]{0.3\textwidth}
\includegraphics[width=\textwidth]{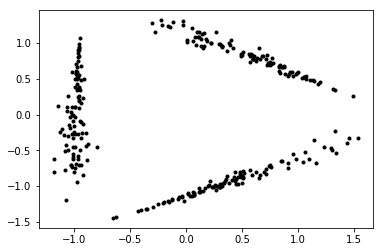} 
        \caption{Radial - 0.05; Tangential - 0.55}
        
    \end{subfigure}
         ~
 \begin{subfigure}[b]{0.3\textwidth}
\includegraphics[width=\textwidth]{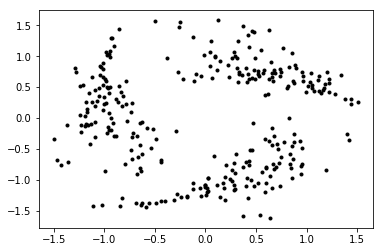} 
        \caption{Radial - 0.25; Tangential - 0.55}
        
    \end{subfigure}
    ~
     \begin{subfigure}[b]{0.3\textwidth}
\includegraphics[width=\textwidth]{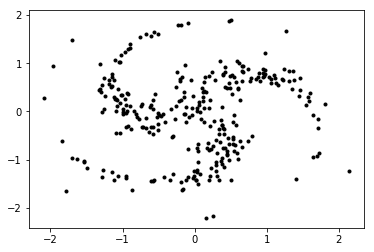} 
        \caption{Radial - 0.55; Tangential - 0.25}
        
    \end{subfigure}
      ~
 \begin{subfigure}[b]{0.3\textwidth}
\includegraphics[width=\textwidth]{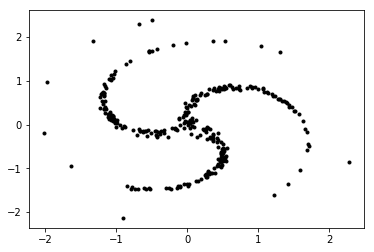} 
        \caption{Radial - 0.55; Tangential - 0.05}
        
    \end{subfigure}
    ~
     \begin{subfigure}[b]{0.3\textwidth}
\includegraphics[width=\textwidth]{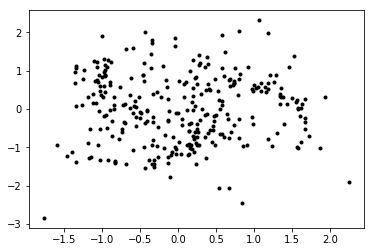} 
        \caption{Radial - 0.55; Tangential - 0.55}
    \end{subfigure}
    
	\caption{Data with completely different shapes can be simulated by varying the tangential and radial parameters of pinwheel dataset.}
	\label{fig:pinwheel}
\end{figure}

\paragraph{Misspecified datasets:}

The pinwheel data is generated by sampling from Gaussian distributions and then stretching and  rotating the data in a controlled manner.  
The centers are equidistant around the unit circle. The variance is controlled by two parameters $r$ and $t$, the radial standard deviation and the tangential standard deviation respectively. The warping is controlled by a third parameter, $s$, the rate parameter.
A datapoint $(x,y)$ belonging to component or arm $k$ is generated as follows:

\begin{align}
    (x',y') & \sim \cN(0,\mathbb{I}) \\
    (x,y) & = ((rx'+1) \cos \theta_k + ty' \sin \theta_k , - (rx'+1) \sin \theta_k + ty' \cos \theta_k), \\
     \text{where} \quad \theta_k & = k\frac{2\pi}{K} + s\times e^{rx'+1} \nonumber 
\end{align}

The MATLAB \footnote{https://github.com/duvenaud/warped-mixtures/blob/master/data/pinwheel.m} and Python\footnote{https://github.com/HIPS/autograd/blob/master/examples/data.py} codes for generating pinwheel datasets are  available online.
We simulated 36 different pinwheel configurations by varying the radial and tangential components for 3-component mixture models using the Autograd package in Python. 
The rate parameter $s$ of warping is fixed at 0.4 for all the 36 different combinations.  This is to test multiple combinations -- clusters with heavy tails to warped mixtures as shown in figure \ref{fig:pinwheel}. The parameters of pinwheel datasets are chosen as follows: 
\begin{itemize}
    \item The radial parameter $r$ is chosen to be one of the six values $\{0.05, 0.15, 0.25, 0.35, 0.45, 0.55\}$
    \item The tangent parameter $t$ is chosen to be one of the six values $\{ 0.05, 0.15, 0.25, 0.35, 0.45, 0.55 \}$
\end{itemize}
By choosing different values of $r$ and $t$, we simulate 36 (6x6) set of pinwheel parameters. For each these 36 pinwheel parameters, we simulate 50 different datasets each with 300 samples, for a total of 1800 datasets.  Please refer to figure \ref{fig:pinwheel} to see the various kinds of component shapes and separations that have been generated.

The tables below show the difference in average ARI between clusterings obtained by EM and those from AD-GD on GMM (table \ref{table:robustness_GMM}) and Pinwheel (table \ref{table:robustness_pinwheel}) datasets. The remaining tables (Tables 22 --29) show the mean and standard deviations over the GMM and Pinwheel datasets for both EM and AD-GD.
From the results in tables \ref{table:robustness_GMM} and \ref{table:robustness_pinwheel}, we observe that in both the cases (misspecification and nomisspecification), EM inference outperforms AG-GD inference. \cite{alexandrovich2014exact}'s work on exact Newton's method using analytical derivatives also report similar results for non misspecified cases.

\begin{table}[!h]
\begin{minipage}{0.5\textwidth}
	\centering
\captionof{table}{Difference in average ARI of GMM datasets}
\label{table:robustness_GMM}

\begin{tabular}{|c||c|c|c|c|c|c|}
    \hline
     $k_{\bmu}$ | $k_{\bSigma}$     & 0.05  & 0.15  & 0.25  & 0.35  & 0.45  & 0.55 \\
    \hline \hline
   0.25  & 0.56  & 0.52  & 0.44  & 0.40  & 0.37  & 0.36 \\
    \hline
    0.5   & 0.43  & 0.55  & 0.49  & 0.44  & 0.41  & 0.40 \\
    \hline
    0.75  & 0.36  & 0.51  & 0.52  & 0.48  & 0.46  & 0.42 \\
    \hline
    1     & 0.36  & 0.48  & 0.55  & 0.53  & 0.50  & 0.50 \\
    \hline
    1.25  & 0.30  & 0.47  & 0.55  & 0.58  & 0.54  & 0.54 \\
    \hline
   1.5   & 0.33  & 0.51  & 0.58  & 0.60  & 0.61  & 0.56 \\
    \hline
    \end{tabular}
    \end{minipage}
\begin{minipage}{0.5\textwidth}

	\centering
\captionof{table}{Difference in average ARI Pinwheel datasets}
\label{table:robustness_pinwheel}

\begin{tabular}{|c||c|c|c|c|c|c|}
    \hline
     $r$ | $t$     & 0.05  & 0.15  & 0.25  & 0.35  & 0.45  & 0.55 \\
    \hline \hline

    0.05  & 0.18  & 0.30  & 0.41  & 0.50  & 0.51  & 0.08 \\
    \hline
    0.15  & 0.47  & 0.50  & 0.56  & 0.58  & 0.57  & 0.24 \\
    \hline
    0.25  & 0.51  & 0.58  & 0.61  & 0.62  & 0.61  & 0.47 \\
    \hline
    0.35  & 0.30  & 0.39  & 0.45  & 0.53  & 0.49  & 0.43 \\
    \hline
    0.45  & 0.12  & 0.17  & 0.20  & 0.24  & 0.27  & 0.30 \\
    \hline
    0.55  & 0.05  & 0.08  & 0.12  & 0.16  & 0.18  & 0.19 \\
    \hline
    \end{tabular}

\end{minipage}

\end{table}

\begin{table}[!h]
\begin{minipage}{0.5\textwidth}
   \centering
\captionof{table}{Average ARI GMM datasets using EM}
\label{table:robustness_GMM_EM}
\begin{tabular}{|c||c|c|c|c|c|c|}
    \hline
     $k_{\bmu}$ | $k_{\bSigma}$     & 0.05  & 0.15  & 0.25  & 0.35  & 0.45  & 0.55 \\
    \hline \hline
    0.25  & 0.78  & 0.63  & 0.52  & 0.46  & 0.42  & 0.41 \\
    \hline
    0.5   & 0.81  & 0.79  & 0.68  & 0.59  & 0.53  & 0.51 \\
    \hline
    0.75  & 0.84  & 0.84  & 0.78  & 0.70  & 0.64  & 0.57 \\
    \hline
    1     & 0.84  & 0.85  & 0.83  & 0.77  & 0.70  & 0.67 \\
    \hline
    1.25  & 0.84  & 0.86  & 0.84  & 0.82  & 0.75  & 0.71 \\
    \hline
    1.5   & 0.83  & 0.86  & 0.86  & 0.83  & 0.81  & 0.74 \\
    \hline
    \end{tabular}
    \end{minipage}
\begin{minipage}{0.5\textwidth}
  \centering
\captionof{table}{Average ARI of GMM datasets using AD-GD}
\label{table:robustness_GMM_Newtons}
\begin{tabular}{|c||c|c|c|c|c|c|}
    \hline
     $k_{\bmu}$ | $k_{\bSigma}$     & 0.05  & 0.15  & 0.25  & 0.35  & 0.45  & 0.55 \\
    \hline \hline
 0.25  & 0.23  & 0.11  & 0.08  & 0.06  & 0.06  & 0.05 \\
    \hline
    0.5   & 0.38  & 0.24  & 0.18  & 0.15  & 0.12  & 0.11 \\
    \hline
    0.75  & 0.47  & 0.33  & 0.26  & 0.21  & 0.18  & 0.16 \\
    \hline
    1     & 0.48  & 0.37  & 0.28  & 0.24  & 0.20  & 0.17 \\
    \hline
    1.25  & 0.54  & 0.39  & 0.29  & 0.24  & 0.21  & 0.17 \\
    \hline
    1.5   & 0.50  & 0.35  & 0.28  & 0.23  & 0.20  & 0.18 \\
    \hline
    \end{tabular}

\end{minipage}
    
\end{table}

\begin{table}[!h]
\begin{minipage}{0.5\textwidth}
   \centering
\captionof{table}{Average ARI of Pinwheel datasets using EM}
\label{table:robustness_GMM_Newtons}

\begin{tabular}{|c||c|c|c|c|c|c|}
    \hline
    $r$ | $t$     & 0.05  & 0.15  & 0.25  & 0.35  & 0.45  & 0.55 \\
    \hline \hline
   0.05  & 0.95  & 1.00  & 1.00  & 1.00  & 0.95  & 0.48 \\
    \hline
    0.15  & 0.95  & 0.97  & 1.00  & 0.98  & 0.93  & 0.58 \\
    \hline
    0.25  & 0.88  & 0.95  & 0.95  & 0.94  & 0.91  & 0.74 \\
    \hline
    0.35  & 0.60  & 0.69  & 0.73  & 0.78  & 0.72  & 0.64 \\
    \hline
    0.45  & 0.36  & 0.41  & 0.41  & 0.44  & 0.45  & 0.46 \\
    \hline
    0.55  & 0.23  & 0.26  & 0.28  & 0.31  & 0.32  & 0.32 \\
    \hline
    \end{tabular}
    \end{minipage}
\begin{minipage}{0.5\textwidth}
  \centering
\captionof{table}{Average ARI Pinwheel datasets using AD-GD}
\label{table:robustness_GMM_EM}

\begin{tabular}{|c||c|c|c|c|c|c|}
    \hline
   $r$ | $t$     & 0.05  & 0.15  & 0.25  & 0.35  & 0.45  & 0.55 \\
    \hline \hline
   0.05  & 0.77  & 0.70  & 0.59  & 0.50  & 0.44  & 0.40 \\
    \hline
    0.15  & 0.48  & 0.47  & 0.44  & 0.40  & 0.36  & 0.34 \\
    \hline
    0.25  & 0.37  & 0.37  & 0.34  & 0.32  & 0.29  & 0.27 \\
    \hline
    0.35  & 0.30  & 0.30  & 0.28  & 0.25  & 0.23  & 0.21 \\
    \hline
    0.45  & 0.24  & 0.24  & 0.22  & 0.20  & 0.18  & 0.16 \\
    \hline
    0.55  & 0.19  & 0.18  & 0.17  & 0.15  & 0.14  & 0.12 \\
    \hline
    \end{tabular}

\end{minipage}
\end{table}

\begin{table}[!h]
\begin{minipage}{0.5\textwidth}
   \centering
\captionof{table}{Std. Deviation of ARI GMM datasets using EM}
\label{table:robustness_GMM_EM_std}
\begin{tabular}{|c||c|c|c|c|c|c|}
    \hline
     $k_{\bmu}$ | $k_{\bSigma}$     & 0.05  & 0.15  & 0.25  & 0.35  & 0.45  & 0.55 \\
    \hline \hline
  0.25  & 0.30  & 0.27  & 0.25  & 0.24  & 0.22  & 0.23 \\
    \hline
    0.5   & 0.29  & 0.25  & 0.26  & 0.27  & 0.26  & 0.25 \\
    \hline
    0.75  & 0.26  & 0.23  & 0.25  & 0.26  & 0.26  & 0.27 \\
    \hline
    1     & 0.25  & 0.23  & 0.22  & 0.26  & 0.26  & 0.26 \\
    \hline
    1.25  & 0.25  & 0.24  & 0.23  & 0.22  & 0.26  & 0.26 \\
    \hline
    1.5   & 0.24  & 0.23  & 0.22  & 0.23  & 0.22  & 0.26 \\
    \hline
    \end{tabular}
    \end{minipage}
\begin{minipage}{0.5\textwidth}
  \centering
\captionof{table}{Std. Deviation of  ARI of GMM datasets using AD-GD}
\label{table:robustness_GMM_Newtons_std}
\begin{tabular}{|c||c|c|c|c|c|c|}
    \hline
     $k_{\bmu}$ | $k_{\bSigma}$     & 0.05  & 0.15  & 0.25  & 0.35  & 0.45  & 0.55 \\
    \hline \hline
  0.25  & 0.21  & 0.13  & 0.09  & 0.08  & 0.07  & 0.05 \\
    \hline
    0.5   & 0.26  & 0.21  & 0.17  & 0.15  & 0.12  & 0.10 \\
    \hline
    0.75  & 0.25  & 0.23  & 0.21  & 0.19  & 0.16  & 0.14 \\
    \hline
    1     & 0.28  & 0.23  & 0.21  & 0.19  & 0.17  & 0.15 \\
    \hline
    1.25  & 0.29  & 0.24  & 0.21  & 0.18  & 0.16  & 0.15 \\
    \hline
    1.5   & 0.27  & 0.25  & 0.23  & 0.20  & 0.18  & 0.16 \\
    \hline
    \end{tabular}

\end{minipage}
    
\end{table}

\begin{table}[!h]
\begin{minipage}{0.5\textwidth}
   \centering
\captionof{table}{Std. Deviation of ARI Pinwheel datasets using EM}
\label{table:robustness_pinwheel_EM_std}

\begin{tabular}{|c||c|c|c|c|c|c|}
    \hline
    $r$ | $t$      & 0.05  & 0.15  & 0.25  & 0.35  & 0.45  & 0.55 \\
    \hline \hline
 0.05  & 0.15  & 0.00  & 0.00  & 0.00  & 0.16  & 0.32 \\
    \hline
    0.15  & 0.15  & 0.12  & 0.00  & 0.08  & 0.15  & 0.34 \\
    \hline
    0.25  & 0.12  & 0.10  & 0.10  & 0.08  & 0.06  & 0.25 \\
    \hline
    0.35  & 0.15  & 0.16  & 0.17  & 0.13  & 0.18  & 0.17 \\
    \hline
    0.45  & 0.13  & 0.12  & 0.16  & 0.19  & 0.18  & 0.16 \\
    \hline
    0.55  & 0.08  & 0.08  & 0.09  & 0.11  & 0.12  & 0.11 \\
    \hline
    \end{tabular}
    \end{minipage}
\begin{minipage}{0.5\textwidth}
   \centering
\captionof{table}{Std. Deviation of  ARI of Pinwheel datasets using AD-GD}
\label{table:robustness_pinwheel_Newtons_std}
\begin{tabular}{|c||c|c|c|c|c|c|}
    \hline
     $r$ | $t$      & 0.05  & 0.15  & 0.25  & 0.35  & 0.45  & 0.55 \\
    \hline \hline
0.05  & 0.03  & 0.04  & 0.04  & 0.04  & 0.05  & 0.05 \\
    \hline
    0.15  & 0.04  & 0.04  & 0.05  & 0.04  & 0.05  & 0.04 \\
    \hline
    0.25  & 0.04  & 0.04  & 0.04  & 0.04  & 0.04  & 0.04 \\
    \hline
    0.35  & 0.03  & 0.03  & 0.04  & 0.04  & 0.03  & 0.03 \\
    \hline
    0.45  & 0.03  & 0.03  & 0.03  & 0.03  & 0.03  & 0.03 \\
    \hline
    0.55  & 0.03  & 0.03  & 0.03  & 0.03  & 0.03  & 0.03 \\
    \hline
    \end{tabular}
    \end{minipage}
\end{table}

\subsection{Noise}
\label{section:appendix_noise}

\begin{figure*}[h]
\centering
 \begin{subfigure}[b]{0.19\textwidth}
\includegraphics[width=\textwidth]{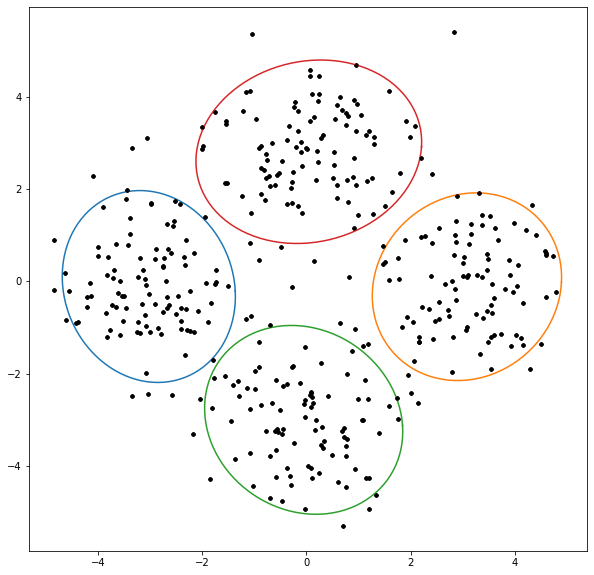} 
        \caption{\# of noise samples = 10}
        
    \end{subfigure}
         ~
 \begin{subfigure}[b]{0.19\textwidth}
\includegraphics[width=\textwidth]{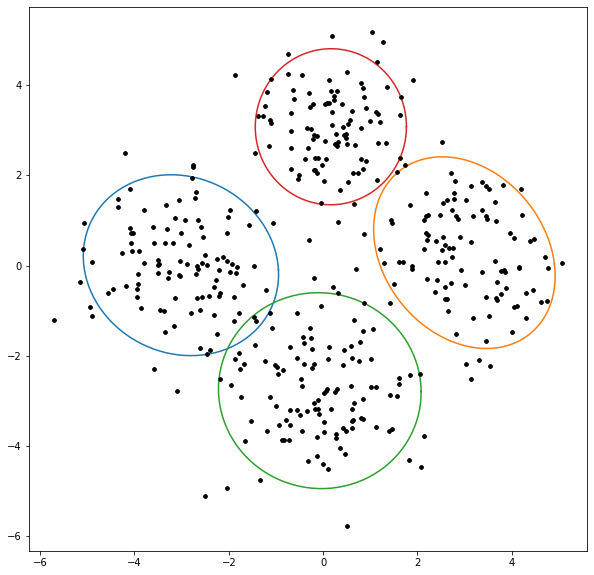} 
        \caption{\# of noise samples = 20}
        
    \end{subfigure}
    ~
     \begin{subfigure}[b]{0.19\textwidth}
\includegraphics[width=\textwidth]{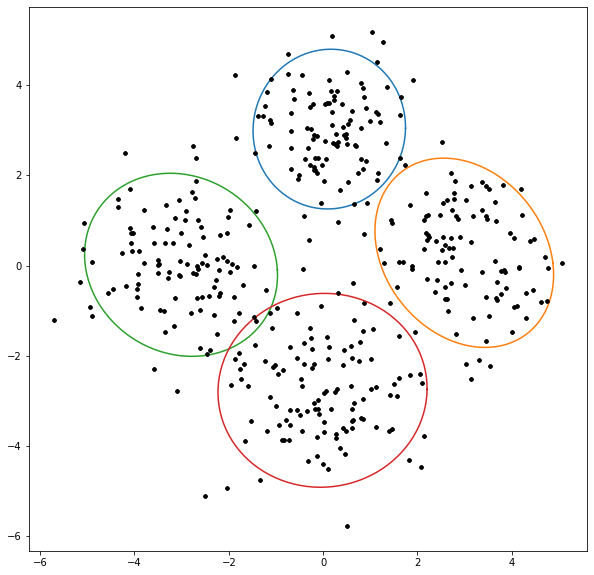} 
        \caption{\# of noise samples = 30}
        
    \end{subfigure}
      ~
 \begin{subfigure}[b]{0.19\textwidth}
\includegraphics[width=\textwidth]{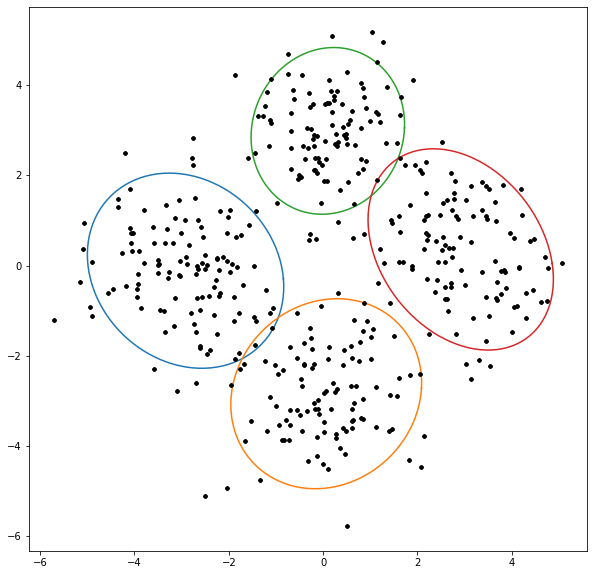} 
        \caption{\# of noise samples = 40}
        
    \end{subfigure}
    ~
     \begin{subfigure}[b]{0.19\textwidth}
\includegraphics[width=\textwidth]{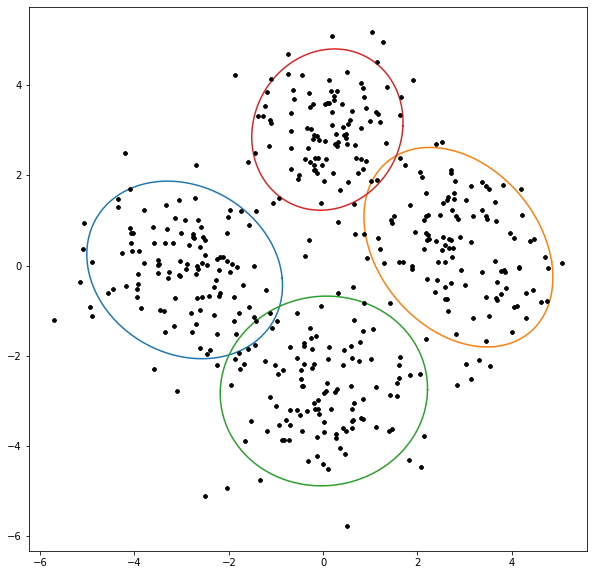} 
        \caption{\# of noise samples = 50}
    \end{subfigure}
    
	\caption{ As the number of noisy samples increases, the affect of misspecification also increases}
	\label{fig:uniform_noise_not_well_separated}
\end{figure*}

\begin{figure*}[h]
\centering
 \begin{subfigure}[b]{0.19\textwidth}
\includegraphics[width=\textwidth]{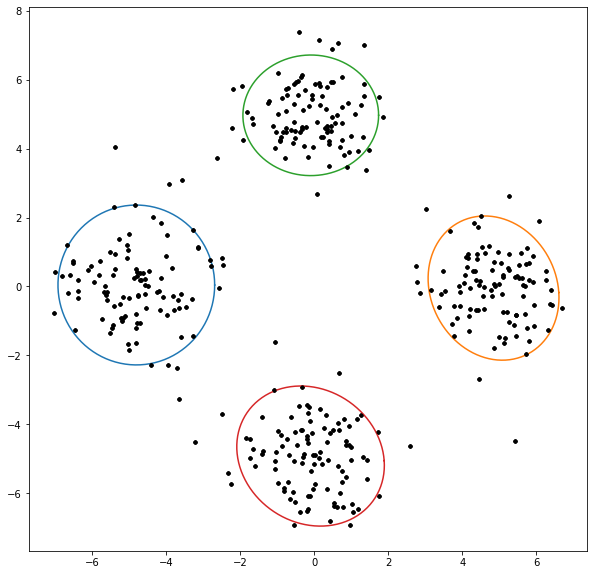} 
        \caption{\# of noise samples = 10}
        
    \end{subfigure}
         ~
 \begin{subfigure}[b]{0.19\textwidth}
\includegraphics[width=\textwidth]{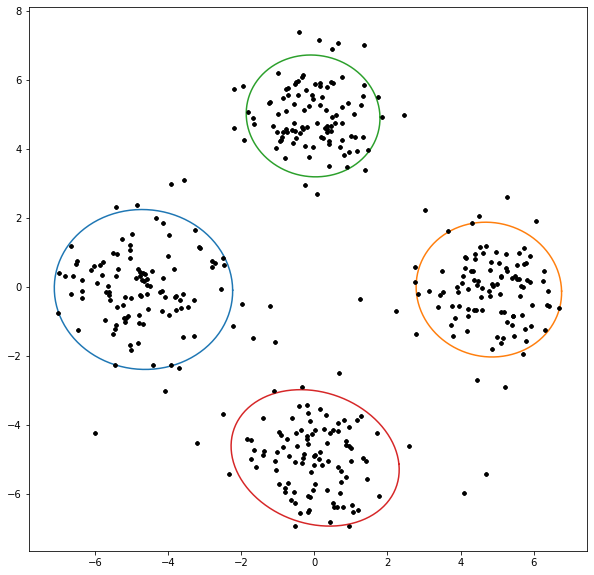} 
        \caption{\# of noise samples = 20}
        
    \end{subfigure}
    ~
     \begin{subfigure}[b]{0.19\textwidth}
\includegraphics[width=\textwidth]{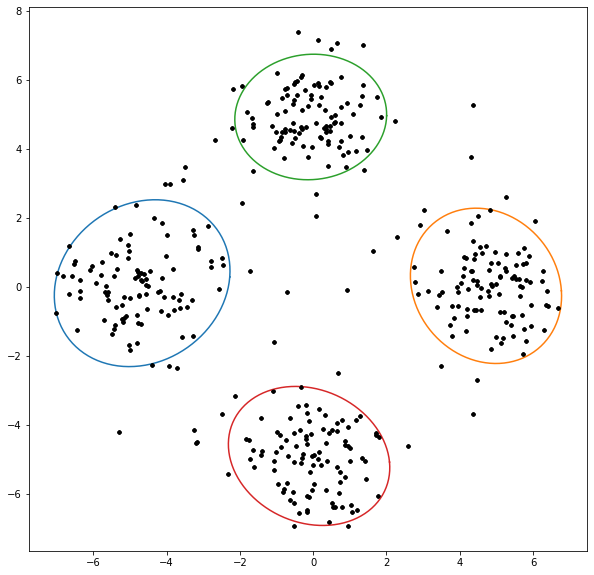} 
        \caption{\# of noise samples = 30}
        
    \end{subfigure}
      ~
 \begin{subfigure}[b]{0.19\textwidth}
\includegraphics[width=\textwidth]{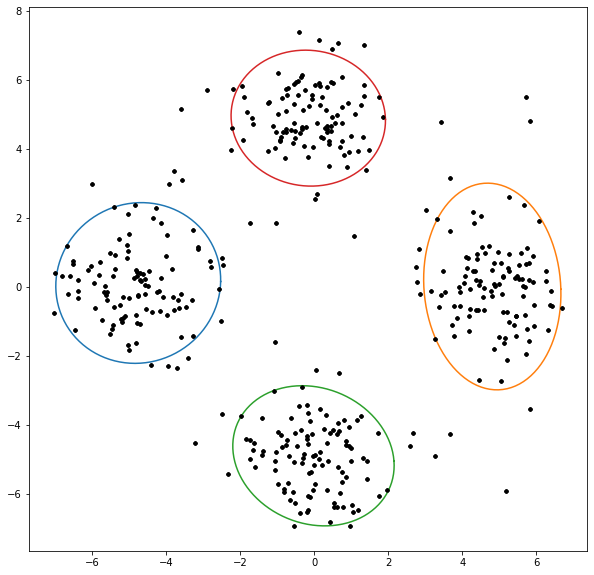} 
        \caption{\# of noise samples = 40}
        
    \end{subfigure}
    ~
     \begin{subfigure}[b]{0.19\textwidth}
\includegraphics[width=\textwidth]{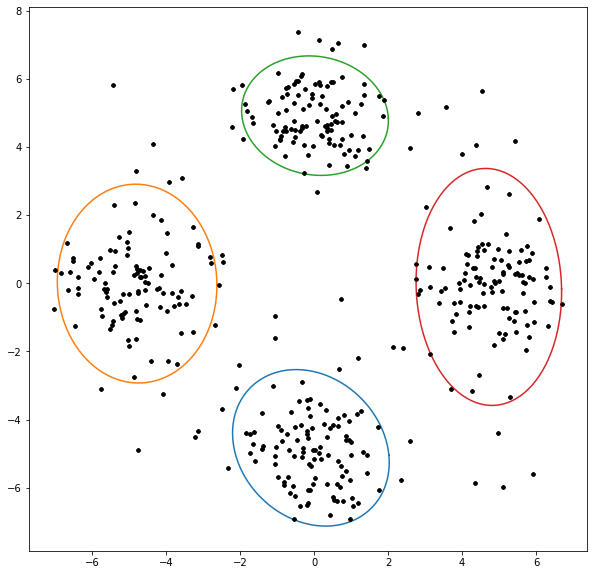} 
        \caption{\# of noise samples = 50}
    \end{subfigure}
    
	\caption{ Increasing the cluster separation mitigates the effect of uniform noise on specification }
	\label{fig:uniform_noise_well_separated}
\end{figure*}

\begin{figure*}[h]
\centering
 \begin{subfigure}[b]{0.19\textwidth}
\includegraphics[width=\textwidth]{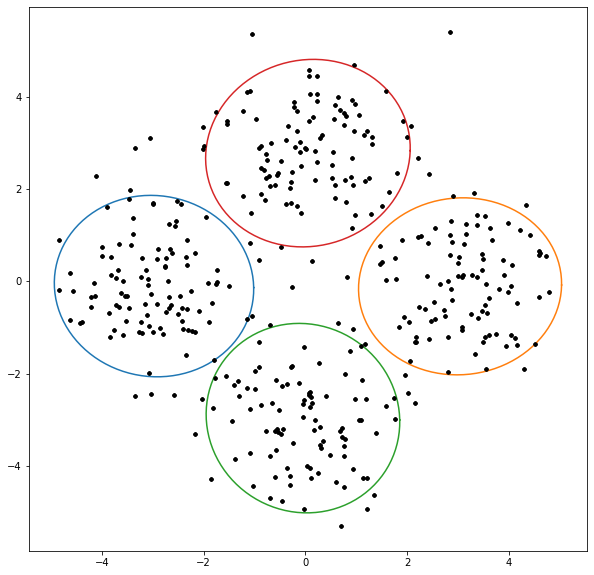} 
        \caption{\# of noise samples = 10}
        
    \end{subfigure}
         ~
 \begin{subfigure}[b]{0.19\textwidth}
\includegraphics[width=\textwidth]{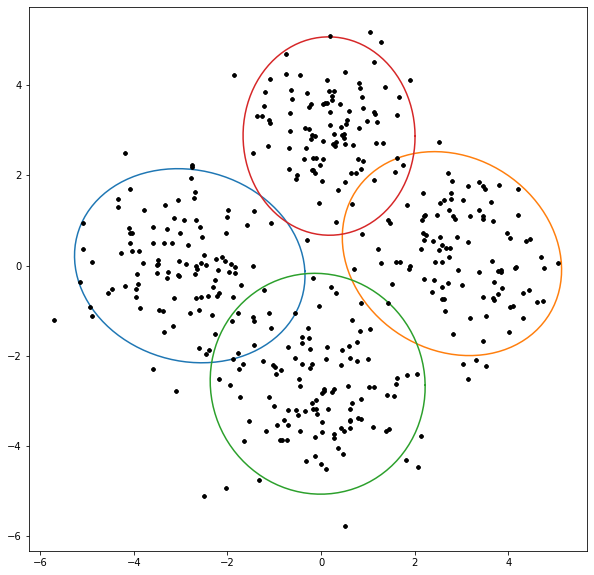} 
        \caption{\# of noise samples = 20}
        
    \end{subfigure}
    ~
     \begin{subfigure}[b]{0.19\textwidth}
\includegraphics[width=\textwidth]{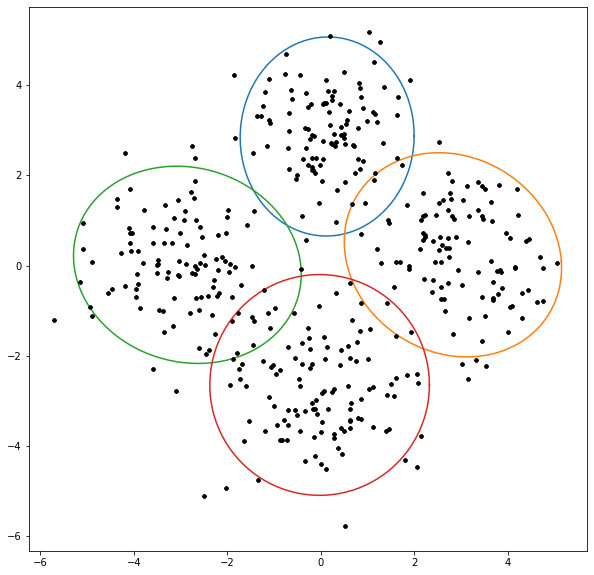} 
        \caption{\# of noise samples = 30}
        
    \end{subfigure}
      ~
 \begin{subfigure}[b]{0.19\textwidth}
\includegraphics[width=\textwidth]{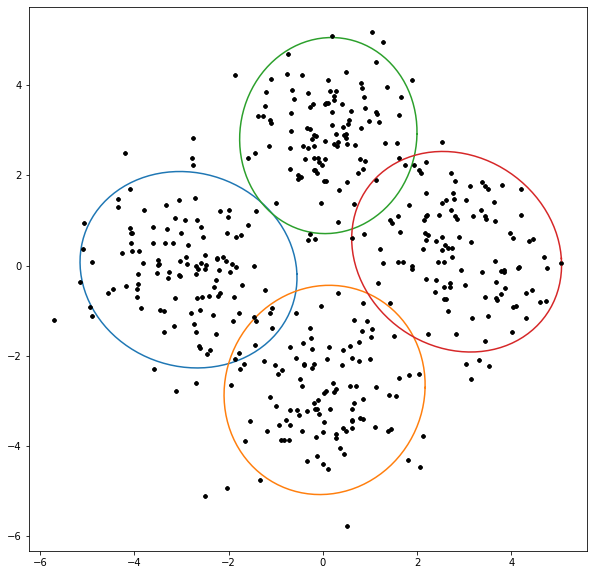} 
        \caption{\# of noise samples = 40}
        
    \end{subfigure}
    ~
     \begin{subfigure}[b]{0.19\textwidth}
\includegraphics[width=\textwidth]{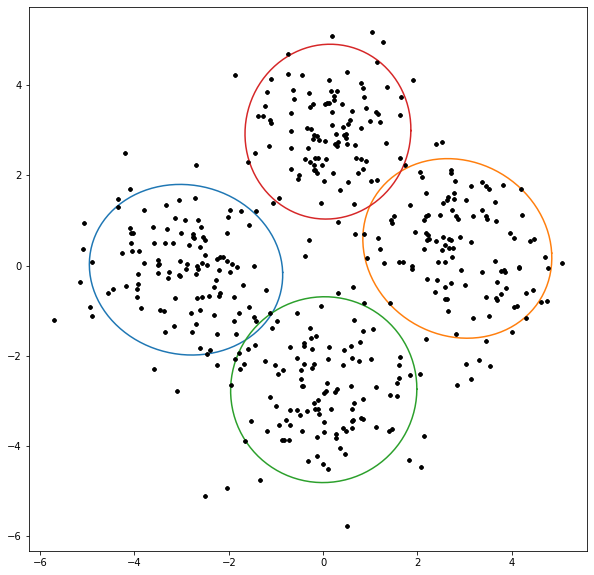} 
        \caption{\# of noise samples = 50}
    \end{subfigure}
    
	\caption{ SIA improves the clustering performance in the uniform noise case}
	\label{fig:uniform_noise_not_well_separated_sia_improv}
\end{figure*}

We construct a four component GMM with unit covariances and means $\{ (-3,0), (0,3), (0,-3), (3,0) \}$. We sample 50 datapoints from each component. We add datapoints sampled uniformly from the square ${(-6,6) \times (-6,6)}$ as noise. 
We fit a GMM using EM in each case.
Fig. \ref{fig:uniform_noise_not_well_separated} shows that as we increase the number of noisy datapoints added to the our original data sampled from GMM, the adverse effect of misspecification on EM increases.  
Fig. \ref{fig:uniform_noise_not_well_separated_sia_improv}
shows that SIA performs well and obtains good clustering.
Table \ref{table:noise_sia_improv} shows the loglikelihood, KLF, KLB, MPKL and ARI values for each case, before and after step 2 of SIA.
We repeat the same experiment of adding noisy data samples, but with increased cluster separation by choosing the means as $\{ (-5,0), (0,5), (0,-5), (5,0) \}$. Fig. \ref{fig:uniform_noise_well_separated} shows that increasing the cluster separation mitigates the effect of misspecification with EM.

\begin{table*}[!h]
\centering
\captionof{table}{ Improvement in clustering performance before and after SIA step-2 with increasing noisy samples $(10, \ldots, 50)$}
\label{table:noise_sia_improv}

\begin{tabular}{ c|p{1.35cm} p{1.25cm}p{1.25cm}p{1.25cm}p{1.25cm}|p{1.35cm}p{1.25cm}p{1.25cm}p{1.25cm}p{1.25cm} } 
 \hline
  & \multicolumn{5}{c|}{Before SIA}  & \multicolumn{5}{c}{After SIA}   \\ \hline
 \# of samples & LL & KLF & KLB & MPKL & ARI  & LL & KLF & KLB & MPKL & ARI \\ \hline %
 
 10 & -1676.9 & 75.6 & 85.2 & 4.4 & 0.91 & -1681.3 & 69.0 & 70.6 & 1.2 & 0.92 \\ %
 20 & -1719.6 & 85.2 & 68.2 & 8.4 & 0.87 & -1737.8 & 51.3 & 45.0 & 2.4 & 0.89 \\ %
 30 & -1761.1 & 64.0 & 87.2 & 6.9 & 0.88 & -1778.4 & 44.2 & 52.8 & 2.3 & 0.89 \\ %
 40 & -1806.2 & 78.1 & 67.3 & 5.6 & 0.87 & -1815.9 & 55.2 & 51.7 & 2.2 & 0.89 \\ %
 50 & -1839.6 & 79.4 & 67.1 & 6.2 & 0.87 & -1845.5 & 70.2 & 65.7 & 2.0 & 0.88 \\ \hline 

\end{tabular}
\end{table*}

\subsection{Wallclock time}
\label{app:complexity}
We benchmark the runtime of SIA with respect to AD-GD and EM. We do so by simulating 50 datasets, each containing 100 datapoints sampled randomly, for each setting. 
We run all the algorithms to convergence in likelihood (tolerance 1e-5) or a maximum of 50 iterations, which ever happens earlier. 
We evaluate with the number of clusters $K = \{3,5\}$ and data dimensions $p=\{ 2,5,10,50 \}$. 
All experiments were run using Autograd, Numpy and Sklearn packages in Python 3.7 on Dell Windows 10 machine (Intel i7-6700 quadcore CPU@3.40GHz; 8 GB RAM; 500 GB HDD). 

Table \ref{table:runtime} shows the runtime of all the algorithms.
We observed that with increasing dimensionality, the number of iterations for EM reduced, and at $p=50$, EM failed to go beyond iteration 1. 
As expected, SIA takes roughly double the time taken by AD-GD.

\begin{table}[!h]
\centering
\captionof{table}{Average Runtime in seconds with data dimensions $2,5,10,50$.}
\label{table:runtime}
\begin{tabular}{c c p{1cm}p{1cm}p{1cm}p{1cm} } 

 \hline
 \# of clusters& Algorithm &  2 &  5 & 10  & 50   \\ \hline
3 & SIA &  7.900  & 7.926 & 8.037 & 11.508\\ %
3 & AD-GD & 4.471 & 4.467 & 4.529 & 6.50   \\ %
3 & EM & 0.009 & 0.006 & 0.009 & - \\ \hline %
 5 &  SIA & 20.571 & 20.785  & 21.238  & 30.041 \\ %
 5 &  AD-GD &  11.127 & 11.194  & 11.433  & 16.167 \\ %
 5 & EM & 0.036 &  0.025 &  0.014 &  - \\ \hline
 
\end{tabular}
\end{table}

\subsection{Summary statistics of solutions in sets 1--4 for EM}
\label{appendix:binning}

\begin{table}[H]
\centering
\captionof{table}{ Summary statistics of EM clustering solutions over 100 random initializations on the Pinwheel dataset (shown in fig. \ref{fig:SOGD_manifold}), grouped into 4 sets based on AIC ranges. Mean and standard deviation of AIC, ARI, component weights and covariance determinants, over solutions in each set. }
\label{table:binning_em}

\begin{tabular}{ cc|p{1.35cm} p{1.25cm}p{1.25cm}p{1.25cm}p{1.25cm}p{1.35cm}p{1.25cm}p{1.25cm} } 
 \hline
Set & AIC Range & AIC & ARI & $\pi_1$ & $\pi_2$ & $\pi_3$  & $|\bSigma_1|$ & $|\bSigma_2|$ & $|\bSigma_3|$  \\ \hline 
1 & 771-773 & 771.9 \newline (0.0005) & 0.625 \newline (3e-16) & 0.257 \newline(5e-4) & 0.265 \newline(7e-5) & 0.477 \newline(1e-4) & 0.0002 \newline (4e-7) & 0.0005 \newline (1e-7) & 0.125 \newline (1e-4) \\ \hline 
2 & 781-782 & 781.1 \newline (2e-6) & 0.912 \newline (0) & 0.306 \newline (4e-6) & 0.341 \newline (3e-6) & 0.352 \newline (7e-6) & 7e-4 \newline(1e-7) & 0.01 \newline(4e-7) & 0.01 \newline(2e-6) \\ \hline 
3 & 786-787 & 786.8 \newline(0.001) & 0.652 \newline (0.002) & 0.257 \newline (5e-5) & 0.266 \newline (4e-4) & 0.475 \newline (4e-4) & 2e-4 \newline 2e-7 & 5e-4 \newline (6e-6)  & 0.157 \newline (5e-4) \\ \hline 
4 & 788-850 & 810.7 \newline(10.84)  & 0.840 \newline(0.084) & 0.287 \newline (0.022) & 0.319 \newline (0.009) & 0.393  \newline (0.03) & 3e-3 \newline (1e-3) & 0.04 \newline(0.02)  & 0.84 \newline(0.084)  \\ \hline 
 
\end{tabular}
\end{table}

\subsection{Illustration of unbalanced clusters on Pinwheel data}
\label{app:unbalanced}
\begin{figure}[h!]
\centering
\captionsetup{justification=centering}
    \begin{subfigure}[b]{0.2 \textwidth}
\includegraphics[width= \textwidth]{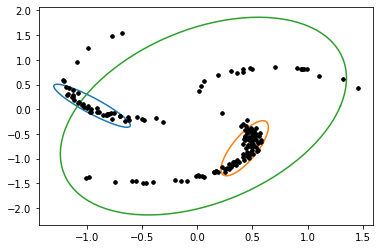}
        \caption{After Step 1: ARI = 0.602\\ MPKL = 220.887 \\ Loglikelihood = -139.4}
        
    \end{subfigure}%
    ~
 \begin{subfigure}[b]{0.2 \textwidth}
\includegraphics[width= \textwidth]{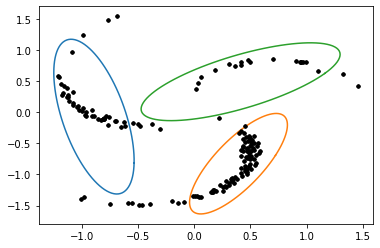}
        \caption{After Step 2: ARI = 0.804 \\  MPKL  = 41.77, \\ Loglikelihood = -191.9 }
        
    \end{subfigure}
    \vrule
    \begin{subfigure}[b]{0.2\textwidth}
\includegraphics[width= \textwidth]{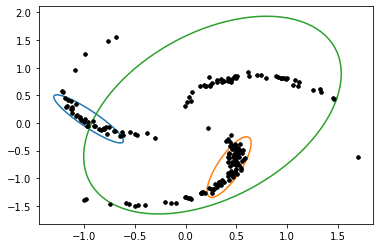}
        \caption{After Step 1: ARI = 0.606 \\ MPKL = 279.769,\\ Loglikelihood = -228.2}
    \end{subfigure}%
    ~
 \begin{subfigure}[b]{0.2\textwidth}
\includegraphics[width= \textwidth]{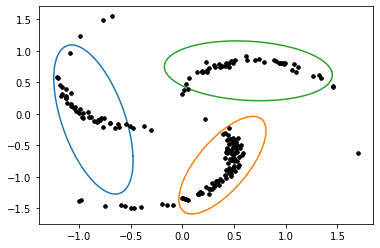}
        \caption{After Step 2:  ARI =0.847  \\  MPKL  = 39.251, \\ Loglikelihood = -261.52 }
        
    \end{subfigure}
	\caption{Clustering using SIA for unbalanced pinwheel data. Componentwise number of datapoints:\\ $\{100,50,20\}$ (a and b), $\{100,50,50\}$ (c and d).}
	\label{fig:sia_unbalanced}
\end{figure}

We illustrate the effect of unbalanced clusters on the same pinwheel data that is discussed in section \ref{sec:misspec}.
In the first experiment, instead of (100,100,100) number of points respectively in the three clusters, we have (100,50,20) number of points respectively for the three clusters. Please refer to fig. \ref{fig:sia_unbalanced} (a) and (b) for the clustering results with SIA with this dataset. 
In the second experiment we have (100,50,50) number of points. Please refer to fig. \ref{fig:sia_unbalanced} (c) and (d) for the results obtained from SIA. %

\section{Closed Forms}
\label{appendix:closed_forms}
In this section, we illustrate the difficulty of deriving the closed form expressions of the gradients of the penalized likelihood used in SIA.
These closed form expressions are required to develop an EM algorithm for inference.
Consider the mean update in EM for the unpenalized loglikelihood:

\begin{align}
    \frac{\partial \log \mathcal{L} }{\partial \bmu_k} = - \sum_{i = 1}^n \gamma(z_{ik}) \bSigma_k^{-1} (\bx_i - \bmu_k)
    \label{eqn:unpenalized_meanupdate}
\end{align}
where $\gamma(z_{ik})$ is defined as the responsibility and is equal to $\frac{\pi_{k} \mathcal{N}\left(\mathbf{x}_{i} | \boldsymbol{\mu}_{k}, \mathbf{\Sigma}_{k}\right)}{\sum_{j=1}^{K} \pi_{j} \mathcal{N}\left(\mathbf{x}_{i} | \boldsymbol{\mu}_{j}, \mathbf{\Sigma}_{j}\right)}$.  %
Setting $\frac{\partial \log \mathcal{L} }{\partial \bmu_k}$
to zero, we obtain $\bmu_k = \frac{\sum_{i=1}^{n} \gamma(z_{ik}) \bx_i  }{ \sum_{i=1}^{n} \gamma(z_{ik}) }$. Note that the computation of the mean update $\bmu_k$ can be easily parallelized in the M-step because it does not involve other mean terms $\bmu_j$ where $j \ne k$.

Now, consider the gradient of the penalized loglikelihood:

\begin{align}
    \begin{split}
       \\ \frac{\partial \log (\mathcal{L} - w_1 \times KLF - w_2  \times KLB ) }{\partial \bmu_k} =
\\ - \sum_{i = 1}^n \gamma(z_{ik}) \bSigma_k^{-1} (\bx_i - \bmu_k) - w_1 \sum_{j \ne k} 2\bSigma_j^{-1} (\bmu_k - \bmu_j) 
 -w_2 \sum_{j \ne k}  2\bSigma_k^{-1} (\bmu_k - \bmu_j) 
    \end{split}
\end{align}

Setting the above expression to zero, we can obtain the expression for the mean $\bmu_k$ as:

\begin{align}
    \begin{split}
        \bmu_k = \left( \sum_{i = 1}^n \gamma(z_{ik}) \bSigma_k^{-1} - (w_1 + w_2) ( \sum_{j \ne k} \left( \bSigma_j^{-1} + \bSigma_k^{-1} \right) ) \right) ^{-1} \\
  \left(  \sum_{i=1}^{n} \gamma(z_{ik}) \bSigma_k^{-1} \bx_i  -  (w_1 + w_2) ( \sum_{j \ne k} \left( \bSigma_j^{-1} + \bSigma_k^{-1} \right) \bmu_j )  \right)   
    \end{split}
\end{align}

As seen above, 
even though the process of obtaining the updates is straightforward, it is laborious. 
For the update of covariance estimates, we could not find a closed form estimate for the penalized loglikelihood. 
Further, note the mean update requires knowledge of other mean terms $\bmu_j$ and hence cannot be parallelized. 
EM based updates for the more flexible models such as MFA will be even more cumbersome.

In contrast, using Gradient Descent (GD)  based inference using Automatic Differentiation (AD) avoids the need to compute such laborious gradient updates by hand or to hard code them in the software program. 
AD-GD based inference approach is model-agnostic as AD software are blackbox tools where we just need to input the loglikelihood; they don't require closed forms of the gradients.
Thus the same software implementation can be easily extended for more flexible models such as MFA, PGMM etc. This is not the case with the traditional EM based inference where the updates have to be derived and re-implemented for each model.

\section{Theoretical Results}
\subsection{Proof of Theorem 1}
\label{app:proof_lemma1}

{First, we show the boundedness for the simple case of a univariate mixture model as it is straightforward to follow and later proceed to the proof of the multivariate case. Throughout the proof we omit the superscript $`t'$ on the estimates for ease of mathematical exposition.  
}

\begin{proof}

\textbf{Univariate:} With loss of generality, let the component 1 - $\mathcal{N}(x; \mu_1, \sigma_1^2)$ - is the one that is fitting over a single point and reducing its variance $\hat \sigma_1^2 \to 0$; also, assume that $\hat \sigma_1^2 \leq \hat \sigma_2^2 \leq c < \infty$

First note that the value of likelihood satisfies the following inequality:
\begin{align}
 \mathcal{N}(x;\mu, \sigma^2) \leq \frac{1}{\sqrt{2\pi} \sigma }   
 \label{eq:normalityineq}
\end{align}

For one-dimensional case, the penalized likelihood of SIA1 for an observation $x_i$ can be written as follows: 

\begin{align}
    \log(\pi_1 \mathcal{N}(x_i; \mu_1, \sigma_1^2) + \pi_2 \mathcal{N}(x_i; \mu_2, \sigma_2^2)) - & w_1\left( \log \frac{\sigma_2}{\sigma_1} + \frac{\sigma_1^2 + (\mu_1 - \mu_2)^2}{2 \sigma_2^2} - \frac{1}{2} \right) - w_2 \left(  \log \frac{\sigma_1}{\sigma_2} + \frac{\sigma_2^2 + (\mu_1 - \mu_2)^2}{2 \sigma_1^2} - \frac{1}{2}  \right) \\
    & \leq \log( \frac{\pi_1}{\sigma_1} + \frac{\pi_2}{\sigma_2}) - w_2 \log \sigma_1 + \frac{w_3}{2 \sigma_1^2} \\
    & \leq \log( \frac{\pi_1 + \pi_2}{\sigma_1} ) - w_2 \log \sigma_1 + \frac{w_3}{2 \sigma_1^2}
\end{align}

\begin{align}
    \log(\hat \pi_1 \mathcal{N}(x_i; \hat \mu_1, \hat \sigma_1^2) + & \hat \pi_2 \mathcal{N}(x_i; \hat \mu_2, \hat \sigma_2^2)) -  w_1\left( \log \frac{\hat \sigma_2}{\hat \sigma_1} + \frac{\hat \sigma_1^2 + (\hat \mu_1 - \hat \mu_2)^2}{2 \hat \sigma_2^2} - \frac{1}{2} \right) - w_2 \left(  \log \frac{\hat \sigma_1}{\hat \sigma_2} + \frac{\hat \sigma_2^2 + (\hat \mu_1 - \hat \mu_2)^2}{2 \hat \sigma_1^2} - \frac{1}{2}  \right) \\ 
    & \text{(By using eq \ref{eq:normalityineq} and setting $(w_3 = w_2 ( \hat \sigma_2^2 + (\hat \mu_1 - \hat \mu_2)^2 ) )$,  and leaving out  -ve terms) } \nonumber \\
    & \leq \log( \frac{\hat \pi_1}{\hat \sigma_1} + \frac{\hat \pi_2}{\hat \sigma_2}) - w_2 \log \hat \sigma_1 - \frac{w_3}{2 \hat \sigma_1^2} - \log \sqrt{2\pi} + w_2 \log c \; \;  \\
    & \leq \log( \frac{\hat \pi_1 + \hat \pi_2}{\hat \sigma_1} ) - w_2 \log \hat \sigma_1 - \frac{w_3}{2 \hat \sigma_1^2} + w_2 \log c   (\text{where $w_4 = 1+ w_2$)} \nonumber     \\
    & \leq - w_4 \log \hat \sigma_1 - \frac{w_3}{2 \hat \sigma_1^2} + w_2 \log c \; \; \\ & \text{(Differentiating wrt $\sigma_1$ and setting to 0, we find $\sigma_1^{*} = \sqrt{\frac{w_3}{w_4}}$}) \;\;  \nonumber \\
    & \leq - w_4 \log \sqrt{\frac{w_3}{w_4}} - 0.5 w_4 + w_2 \log c \label{eq:univariate_bounded}
\end{align}

The terms in equation \ref{eq:univariate_bounded} are all constants hence the SIA 1 likelihood is bounded.  

\textbf{Multivariate:}
$\lambda_1(\hat \Sigma_1), \lambda_1(\hat \Sigma_2) \leq c_1$ where $\lambda_p(.)$ and $\lambda_1(.) $  denotes the smallest and the largest eigenvalue of the $\Bbb R^{p \times p}$ matrix.  The likelihood can be made unbounded by making the determinant of $\hat \Sigma_1$ close to zero i.e. $|\hat \Sigma_1| \to 0$. 

First, we bound the following terms:
\begin{align}
    0.5 w_2 \log (| \hat \Sigma_2 |) & \leq 0.5 w_2 p\log (c_1) = C \text{(some constant)}  
    \label{eq:mvn_ind_bounds}
\end{align}

\begin{align}
    \log(\hat \pi_1 \mathcal{N}(x_i; \hat \mu_1, \hat \Sigma_1) & + \hat \pi_2 \mathcal{N}(x_i; \hat \mu_2, \hat \Sigma_2)) -  0.5* w_1\left( \log \frac{ |\hat \Sigma_2 |}{| \hat \Sigma_1|} + \tr({\hat \Sigma_2^{-1} \hat \Sigma_1}) + (\mu_2-\mu_1)^{T}\hat \Sigma_2^{-1}(\mu_2-\mu_1)^{T} \right) \nonumber \\
    & - 0.5* w_2 \left(  \log \frac{|\hat \Sigma_1|}{|\hat \Sigma_2|} + \tr({\hat \Sigma_1^{-1}\hat \Sigma_2}) + (\mu_2-\mu_1)^{T}\hat \Sigma_1^{-1}(\mu_2-\mu_1)^{T}  \right) \\ 
    & \text{(By using eq \ref{eq:normalityineq} and \ref{eq:mvn_ind_bounds},  and leaving out  -ve terms and absorbing the constants into $C$) } \nonumber\\
    & \leq 0.5 * \log( \frac{\hat \pi_1}{|\hat \Sigma_1|} + \frac{\hat \pi_2}{|\hat \Sigma_2 |}) - 0.5*w_2 \log( |\hat \Sigma_1| )  -  0.5*w_2 \tr{(\hat \Sigma_1^{-1} \hat \Sigma_2)}  \; \;  \\
    & = 0.5* \left(\log( \frac{\hat \pi_1 + \hat \pi_2}{|\hat \Sigma_1|} ) - w_2 \log( |\hat \Sigma_1 |)  -  w_2 \tr{(\hat \Sigma_1^{-1} \hat \Sigma_2)}  \right) \\
    & = 0.5* \left( - \log( {|\hat \Sigma_1|} ) - w_2 \log( |\hat \Sigma_1 |)  - w_2 \tr{(\hat \Sigma_1^{-1} \hat \Sigma_2)}  \right) \\
    & = 0.5* \left( - w_3 \log( {|\hat \Sigma_1|} )   - w_2 \tr{(\hat \Sigma_1^{-1} \hat \Sigma_2)}  \right) \; \; \; \text{(where $w_3 = 1+ w_2$)}  \\
    & = 0.5* \left( w_3 \log( | \hat \Sigma_1^{-1}|) -w_2 \tr{(\hat \Sigma_1^{-1} \hat \Sigma_2)}   \right) \label{eq:concave_bound} 
\end{align}

Note that equation \ref{eq:concave_bound} is concave in $\Sigma_1^{-1}$  as $\log (|.|)$ is a concave function and trace is affine \citep{boyd2004convex}; hence, if a maximum exists, it is the global maximum. We can find the maximum by differentiating with respect to $\Sigma_1^{-1}$ and setting to zero;
\begin{align}
    & \nabla_{\Sigma_1^{-1}} \left( w_3 \log( | \hat \Sigma_1^{-1}|) -w_2 \tr{(\hat \Sigma_1^{-1} \hat \Sigma_2)}   \right) = w_3 \hat \Sigma_1 - w_2 \hat \Sigma_2 = 0 \\
    & \implies \hat \Sigma_1 = \frac{w_2}{w_3} \hat \Sigma_2
\end{align}
Therefore, the final upper bound of the log-likelihood is given by 
\begin{align}
    - 0.5 (1+w_2)\left(  \log (\frac{w_2}{1+w_2} \hat \Sigma_2) + p  \right) + C 
\end{align}
\end{proof}

\section{Likelihood Surface Visualization}
\label{sec:visu}
To understand how the penalization in \eqref{eq:SIA} helps in clustering we visualize the log-likelihood surface using the 
technique of \citet{li2018visualizing} 
for neural network loss landscapes.
Given two sets of parameters $\btheta_1, \btheta_2$, the surface function $ S(\alpha, \beta)=\mathcal{L}\left(\alpha \btheta_1+\beta \btheta_2 \right)$ where $0\le \alpha, \beta \le 1$ is used to analyze the log-likelihood $\mathcal{L}$.

\begin{figure}[h!]
\includegraphics[width=0.45\textwidth]{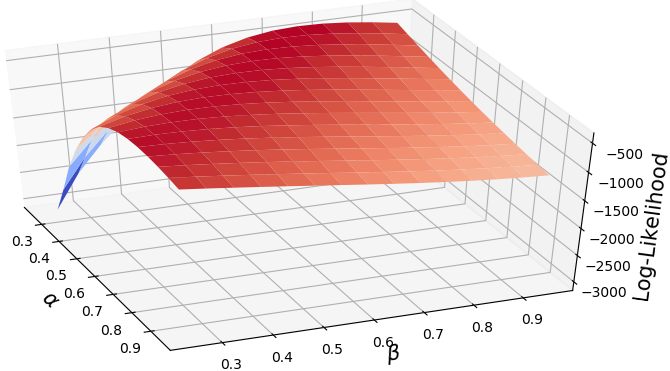}
\includegraphics[width=0.45\textwidth]{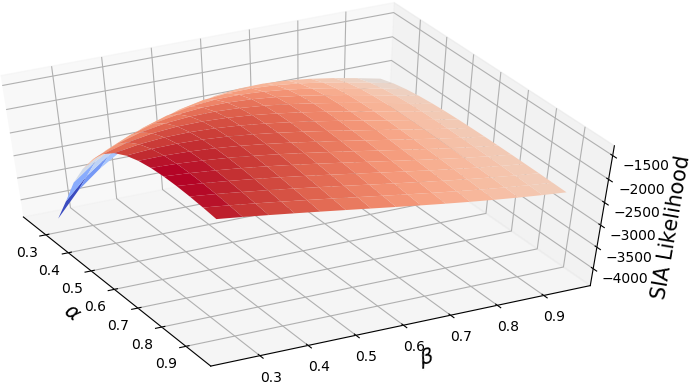}
    \centering
    \caption{ 
    (Left) Log-likelihood shows a plateau region ; (Right) Penalized log-likelihood has no plateau and a clear peak corresponding to good clustering.}
    \label{fig:lossfucn}
\end{figure}

As an illustration, we use 
the Pinwheel data described in section \ref{sec:misspec} 
with the parameters obtained for (g) and (a) of fig. \ref{fig:SOGD_manifold} as $\btheta_1$ and $\btheta_2$ respectively and plot the log-likelihood and penalized log-likelihood in fig. \ref{fig:lossfucn} (a) and (b) respectively.
It is evident
in fig. \ref{fig:lossfucn} (a)
that the  log-likelihood has a significant plateau region 
which has almost similar (unpenalized) likelihood. However, for SIA likelihood in (b), such a plateau does not exist and there exists a clear maxima at $(\alpha = 1, \beta = 0)$. As expected $\btheta_1$ ((g) of fig. \ref{fig:SOGD_manifold}) has a better clustering output, despite having similar (unpenalized) likelihood as that of $\btheta_2$. 

\end{document}